\documentclass[10pt,a4paper]{tencent_lightspeed}
\usepackage{amsmath,amssymb,amsthm}
\usepackage{tabularx,colortbl}
\usepackage{needspace}
\usepackage{algorithm,algorithmic}

\ifPDFTeX
  \newcommand{\tableface}{\fontfamily{ptm}\selectfont}
\else
  \usepackage[no-math]{fontspec}
  \IfFontExistsTF{Times New Roman}
    {\newfontfamily\tableface{Times New Roman}[Ligatures=TeX]}
    {\newcommand{\tableface}{\fontencoding{T1}\fontfamily{ptm}\selectfont}}

  \renewcommand{\sfdefault}{phv}
  
\fi
\newcommand{\methodname}{LightDecision-1}
\newcommand{\optionSet}{\mathcal{O}}
\newcommand{\targetdist}{\bm{r}}
\newcommand{\decisionmodel}{f_{\bm{\theta}}}
\newcommand{\stopgrad}{\operatorname{sg}}

\definecolor{decisiongold}{HTML}{A18635}

\colorlet{tableheadfg}{black}
\colorlet{tablegroupbg}{lightspeedblue!5}
\colorlet{tableoursbg}{lightspeedblue!15}
\newcommand{\reporttable}{\centering\tableface\footnotesize
  \setlength{\tabcolsep}{2.5pt}\renewcommand{\arraystretch}{1.08}}
\newcommand{\tablehead}[1]{\textcolor{tableheadfg}{\textbf{#1}}}
\newcommand{\bestscore}[1]{\textbf{#1}}
\newcommand{\secondscore}[1]{\underline{#1}}
\newcommand{\ranklegend}{\textbf{Bold} and \underline{underlining} mark the best and second-best values, respectively.}
\newcolumntype{Y}{>{\centering\arraybackslash}X}

\DeclareCaptionFont{tableface}{\tableface}
\title{LightDecision-1: An Open-Vocabulary Probabilistic Multimodal Decision Model}
\author{LightDecision-1 Team}
\reporttype{}
\reportlogo[width=44mm,trim=137bp 136bp 137bp 131bp,clip]{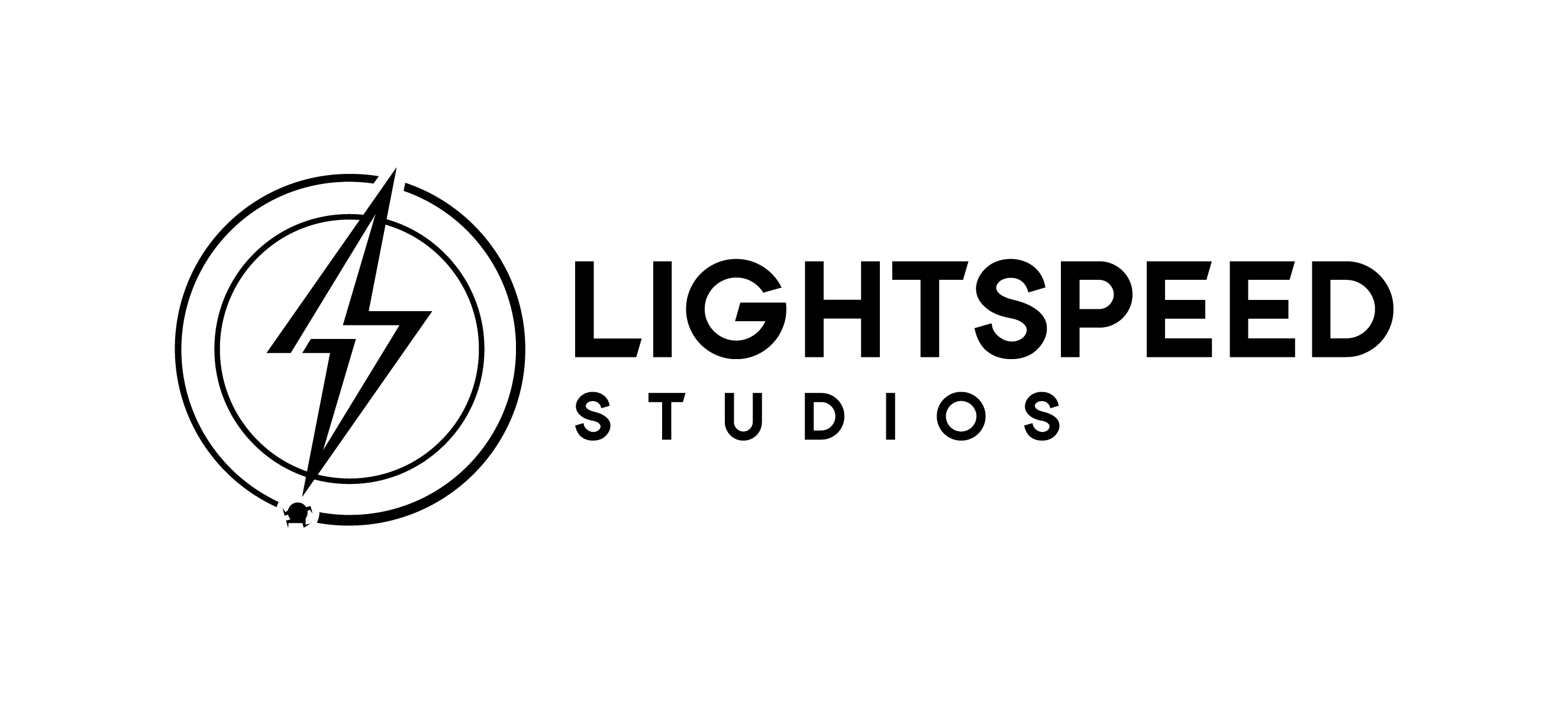}
\reportpartners{%
  \partnerlogo{13.5mm}{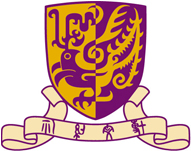}%
  \partnerlogo{13.4mm}{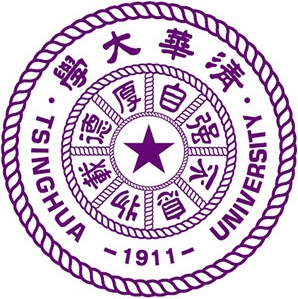}%
  \partnerlogo{14.2mm}{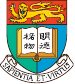}%
  \partnerlogo{13.4mm}{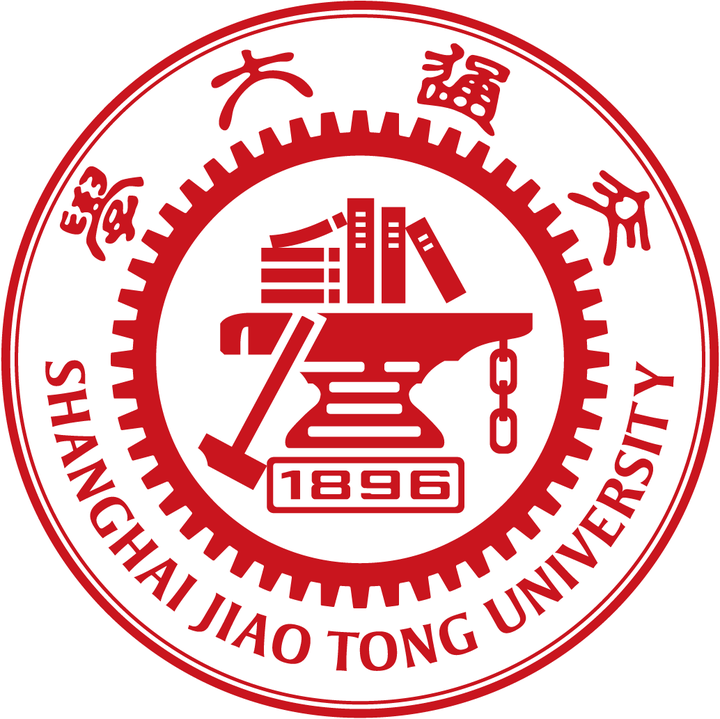}%
  \partnerlogo{13.4mm}{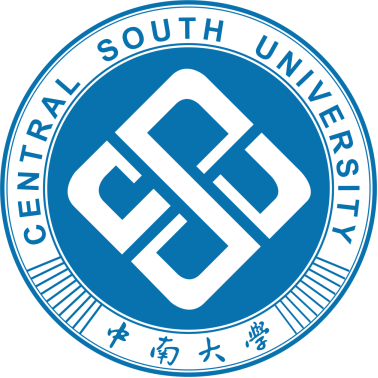}%
}
\reportresources{%
  \href{https://lightdecision.github.io/}{%
    \raisebox{-.22\height}{\includegraphics[height=12pt]{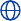}}%
    \hspace{.38em}Project Page}%
  \hspace{1.8em}%
  \href{https://huggingface.co/shenjunhao/lightdecision-1-4b}{%
    \raisebox{-.24\height}{\includegraphics[height=13pt]{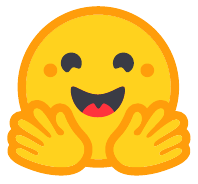}}%
    \hspace{.38em}Hugging Face\hspace{.45em}\textbf{4B}}%
  \hspace{.25em}{\color{black!45}/}\hspace{.25em}%
  \href{https://huggingface.co/shenjunhao/lightdecision-1-27b}{\strut\textbf{27B}}%
  \hspace{1.8em}%
  \href{https://github.com/LightDecision/lightdecision-1}{%
    \raisebox{-.22\height}{\includegraphics[height=12pt]{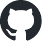}}%
    \hspace{.38em}GitHub}%
}
\abstract{A visual agent acts on what it sees, and needs to observe at each step to make fast and scene-grounded decisions. 
However, existing System-1 decision models risk disrupting visual-semantic representations through reused vocabulary tokens, struggle to capture option interactions without order sensitivity, and scale poorly to large option sets.
Thus, we introduce \textbf{\methodname{}},  a family of 4B and 27B open-vocabulary multimodal System-1 decision models that produce a probability distribution over a variable set of natural-language options in a single forward pass. These options can describe actions or answers.
Specifically, the backbone encodes each option in a separate branch that shares the same multimodal state and query. A permutation-equivariant attention head compares options jointly so that reordering the options preserves the probability assigned to each one.
During the post-training phase, we first apply supervised fine-tuning on hard and soft labels to establish the decision format. Then, we apply reinforcement learning for calibrated decisions with unbiased gradient estimates, sampling multiple options in parallel to improve calibration and training efficiency.
Extensive experiments show strong results at both scales. 
Notably, the 27B model achieves 41.33\% progress on GameWorld and 48.67\% success on EmbodiedBench-Habitat, exceeding the strongest baselines by 5.3 and 1.3 points.
Further analyses confirm sub-second latency, superior calibration, order invariance, action-set scaling, and inference speedup.}
\hypersetup{pdftitle={\titlelist},pdfauthor={LightDecision-1 Team}}

\begin{document}
\maketitle
\vspace{30pt}
\begin{minipage}{\textwidth}
  \centering
  \captionsetup{hypcap=false}
  \includegraphics[width=\linewidth]{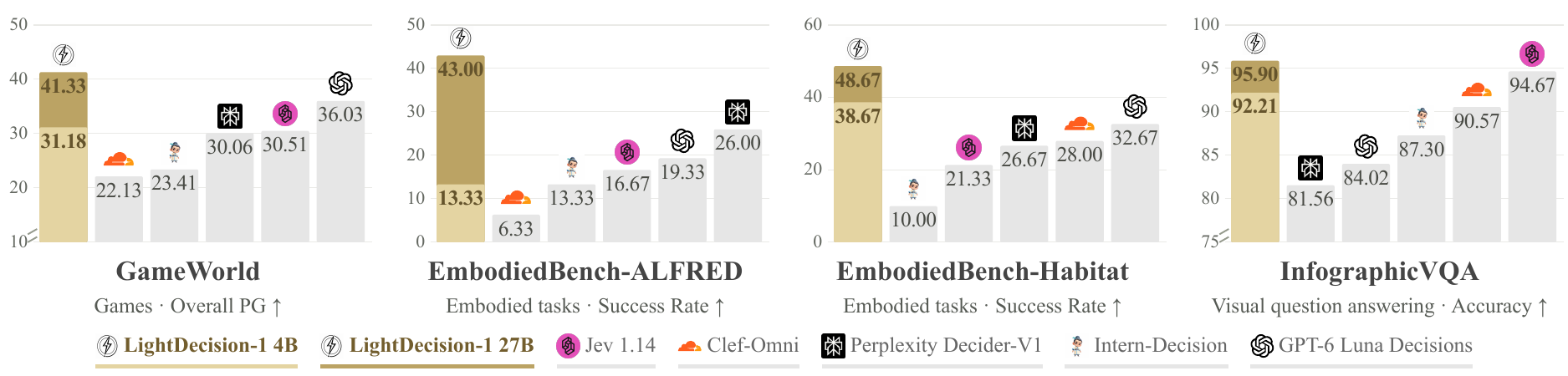}
  \captionof{figure}{\textbf{Visual decision performance across interactive environments and knowledge tasks.} 
  The two gold bars indicate LightDecision-1 4B and 27B, respectively.
  Across all methods, LightDecision-1-27B leads on GameWorld, EmbodiedBench and InfographicVQA.
  Detailed statistics are shown in Table~\ref{tab:vision} .
  }
  \label{fig:overview}
\end{minipage}
\par\medskip

\maintextgeometry
\addtocontents{toc}{\protect\setcounter{tocdepth}{-1}}

\section{Introduction}
\label{sec:introduction}
An interactive agent uses vision to decide how to act, and its actions shape what it sees next. As its backbone, multimodal language models interpret the current scene and generate a response~\citep{liu2023llava,chen2024internvl}. In short, the agent must then map that response to an available action and repeat the process as the scene changes~\citep{koh2024visualwebarena,yao2023react,he2024webvoyager}.
System-1 decision models such as Jev instead assign probabilities to natural-language options in a single forward pass~\citep{typesafe2026jev,microsoft2026decision}.
Recently, Intern-Decisionzk~\citep{zhu2026intern}, Clef-Omni~\citep{cloudflare2026clef} and VideoJev~\citep{zhu2026videojev} have extended direct scoring to images and video.

However, existing methods face three challenges. 
First, they often place all options in a single prompt and reuse pretrained vocabulary tokens for probability prediction~\citep{robinson2023mcqa}, which disrupts pretrained visual-semantic representations and introduces token biases when handling open-vocabulary options~\citep{zheng2024selectors}. 
Second, encoding all options in a single autoregressive sequence introduces order sensitivity, while scoring each option independently prevents the model from comparing candidates and capturing their dependencies (e.g., ``none of these'' in Figure~\ref{fig:architecture}).
Third, interactive environments may offer hundreds of valid actions at each step. Concatenating these options into a long sequence strains context length and dilutes attention~\citep{huang2026threshold}. 

To address these issues, we introduce \textbf{\methodname{}}, a family of 4B and 27B open-vocabulary multimodal System-1 decision models. The backbone encodes the visual state and query once, then reuses that prefix across parallel option branches.
The decision head projects the hidden states into the decision space. Two layers of bidirectional attention let each option draw on the others regardless of order, and a shared bilinear readout scores each option against the query.

To train this probabilistic decision interface, we convert diverse supervision signals into target distributions over options supplied with each query.
We adopt a two-stage post-training strategy. Supervised fine-tuning (SFT) learns from mainly hard labels using cross-entropy, while reinforcement learning for calibrated decisions (RLCD) emphasizes soft labels. RLCD scores all options in a single forward pass and combines exact gradients with unbiased estimates from sampled binary feedback.
Furthermore, \methodname{} processes all option branches in parallel from the cached multimodal prefix to improve inference efficiency.

Extensive experiments show that \methodname{}-27B ranks first on GameWorld, EmbodiedBench and InfographicVQA, exceeding the strongest baselines by 5.3 percentage points in GameWorld progress and 1.3 points in Habitat success.
Among 4B models, \methodname{} leads on GameWorld, Habitat and both VQA benchmarks, outperforming its Qwen3.5-4B backbone across all tasks and even surpassing larger and proprietary decision models on some benchmarks.
Under option permutations, both models exhibit zero selection flips and $<2\times 10^{-7}$ probability drift. For large action sets, \methodname{} generalizes from at most 77 training options to 305-option ALFRED tasks, delivering sub-second P95 latency and $1.60\times$/$2.70\times$ prefix-reuse speedups for an image with 128 options.

Our contributions are threefold:\\
\textbf{(i)} We introduce \methodname{}, an open-vocabulary multimodal decision model combining prefix-shared parallel branches with permutation-equivariant set attention.\\
\textbf{(ii)} We propose a two-stage SFT--RLCD recipe unifying hard and soft supervision, combining analytic gradients with unbiased binary outcome feedback under a strictly proper scoring rule.\\
\textbf{(iii)} Both 4B and 27B models achieve leading interactive and visual performance, combining order invariance, action-set scaling, and sub-second latency.

\section{Related Work}
\label{sec:related-work}

\noindent\textbf{Multimodal Understanding and Reasoning.}
Visual instruction tuning teaches vision-language models to answer questions about objects, attributes, and their relationships~\citep{liu2023llava,dai2023instructblip,chen2024internvl}.
Multimodal reasoning datasets annotate the evidence or intermediate inferences that connect an observation to an answer~\citep{lu2022scienceqa,chen2024m3cot,shao2024visualcot}, while benchmarks evaluate the reasoning and understanding in various senarios~\citep{mmmupro,infographicvqa,chen2026snapbench}.
When the answer turns on a small detail or streaming settings, guided search and visual tools let a model locate, magnify, or mark the relevant region of an image~\citep{wu2024vstar,hu2024visualsketch,shen2025zoomeye,du2026omnistreaming}.
Reinforcement learning for visual reasoning assigns credit at different points in this process, from a correct final answer to an individual reasoning step or a visual description that supports the answer~\citep{wei2025openvisionreasoner,zhang2025r1vl,shen2025sophia}.
\methodname{} answers visual questions in a single forward pass without generating a reasoning trace, and outperforms its backbone on both world knowledge and understanding.

\noindent\textbf{Multimodal Interaction.}
Vision-language models serve both as interactive agents and as backbones for task-specific policies.
As agents, they can answer follow-up questions about an image, issue browser commands from screenshots, or choose keyboard and mouse actions during gameplay~\citep{liu2023llava,he2024webvoyager,tan2025cradle,xu2026workagents}.
As policy backbones, they can be fine-tuned on annotated screenshots to map instructions to click locations~\citep{cheng2024seeclick,lin2025showui} or on robot demonstrations to predict end-effector motion~\citep{kim2025openvla,li2026forgewm}.
\methodname{} is a multimodal System-1 decision model. Rather than generating textual responses or actions, it assigns probabilities to the supplied options given the state and query in a single forward pass.

\noindent\textbf{System-1 Decision Models.}
As aforementioned, System-1 decision models directly predict a distribution over the supplied options.
Jev introduces a typed decision interface~\citep{typesafe2026jev}.
Following this closed-source release, open implementations have explored how to obtain such decisions from language-model backbones~\citep{theoleecj2026semif,li2026llmasjev}, while more proprietary decision models have also emerged~\citep{microsoft2026decision,cloudflare2026clef}.
Recently, this approach has been extended to multimodal inputs. Intern-Decision-4B scores option symbols at masked decision positions~\citep{zhu2026intern}, and VideoJev scores temporal intervals and possible answers in a causal sequence~\citep{zhu2026videojev}.
Meanwhile, new benchmarks have emerged to evaluate both decision accuracy and the quality of predicted probabilities~\citep{mmdm2026,deusser2026evaluating,duan2026jeval}.
However, existing methods risk disrupting visual-semantic representations by reusing pretrained tokens, struggle to capture option interactions without order sensitivity, and face scalability limits when handling large option sets.
Therefore, we introduce \methodname{}, a multimodal System-1 model for open-vocabulary decisions to address these issues.

\section{LightDecision-1}
\label{sec:method}
In a decision instance $x=(s,q,\optionSet)$, the state $s$ provides evidence such as text, images, or video frames. The query $q$ specifies what to decide. We denote the supplied list of $K$ options by $\optionSet=(o_1,\ldots,o_K)$. Their number and vocabulary can vary across instances.
Each $o_i$ has a name and an optional criterion describing its meaning or when it applies. A criterion may refer to another option, so the model evaluates the set jointly given $s$ and $q$.
With trainable parameters $\bm{\theta}\in\mathbb{R}^{M}$, \methodname{} assigns probability $p_i$ to option $o_i$ and returns $\bm{p}=\decisionmodel(x)\in\Delta^{K-1}$, where $\Delta^{K-1}$ is the probability simplex in $\mathbb{R}^{K}$, $p_i\geq 0$, and $\sum_{i=1}^{K}p_i=1$.
The decision head computes the full distribution in a single forward pass without autoregressive generation.

\subsection{Architecture}
\label{sec:decision-head}
\begin{figure*}[t]
  \centering
  \includegraphics[width=\linewidth]{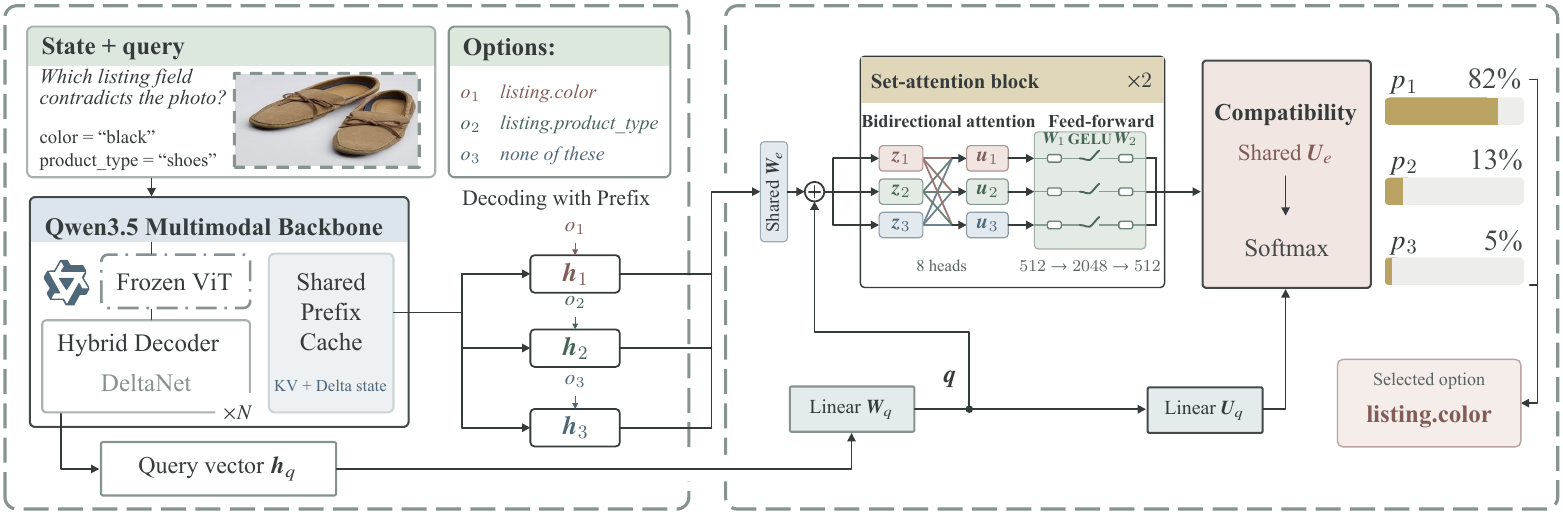}
  \caption{\textbf{LightDecision-1 architecture.} The state (a product photograph) and query form the shared prefix. Each option names a product field or indicates that none applies, and has its own branch. Two set blocks exchange information across options. Within each block, $\bm{z}_i,\bm{u}_i\in\mathbb{R}^{D}$ are the hidden states of option $i$ before and after attention. The feed-forward maps $\bm{W}_1\in\mathbb{R}^{2048\times D}$ and $\bm{W}_2\in\mathbb{R}^{D\times2048}$ act on each row. A query-conditioned readout scores all options and selects the one with the highest probability.}
  \label{fig:architecture}
\end{figure*}

As shown in Figure~\ref{fig:architecture}, we take the decision representation from contextual hidden states of the Qwen3.5 backbone~\citep{qwen2026qwen35}. All options share the same state $s$ and query $q$, which form a common causal prefix with final hidden state $\bm{h}_q\in\mathbb{R}^{H}$, where $H$ is the backbone width. Each option continues from this prefix in its own branch. Its terminal state $\bm{h}_i\in\mathbb{R}^{H}$ encodes the name and criterion of the option.
Then, we project the two kinds of hidden states into a decision space. The projection matrices $\bm{W}_q,\bm{W}_e\in\mathbb{R}^{D\times H}$ and biases $\bm{b}_q,\bm{b}_e\in\mathbb{R}^{D}$ give the query vector $\bm{q}$ and the initial option vector $\bm{z}_i^{(0)}$, both in $\mathbb{R}^{D}$.
\begin{equation}
  \bm{q}=\bm{W}_q\bm{h}_q+\bm{b}_q,\qquad
  \bm{z}_i^{(0)}=\bm{W}_e\bm{h}_i+\bm{b}_e+\bm{q}.
  \label{eq:decision-state}
\end{equation}

If two options are scored independently, adding a third cannot affect their relative scores.
Inspired by Set Transformer\citep{lee2019settransformer}, we use two pre-normalized set-attention blocks to model interactions among options. Each block applies bidirectional attention across all valid options, followed by a feed-forward network shared across options. After the second block, $\bm{z}_i^{(2)}\in\mathbb{R}^{D}$ represents option $i$ in the full set. Finally, a bilinear readout with learned projections $\bm{U}_q,\bm{U}_e\in\mathbb{R}^{D\times D}$ compares this state with the projected query.
It gives each option a logit $\eta_i$ and probability $p_i$.
\begin{equation}
  \eta_i=\frac{(\bm{U}_q\bm{q})^\top\bm{U}_e\bm{z}_i^{(2)}}{\sqrt{D}},
  \qquad p_i=\frac{\exp \eta_i}{\sum_{j=1}^{K}\exp \eta_j}.
  \label{eq:readout}
\end{equation}
This inner product scores each option in the context of the state, query, and full option set. Positional embeddings are not used for options and parameters are shared across options to ensure permutation equivariance, i.e., reordering the options only reorders their probabilities (the formal proof is shown in Appendix~\ref{app:derivations}).
We also project hidden states into a learned decision space rather than directly comparing independently encoded image and option embeddings, since their similarity mainly reflects semantic alignment rather than the query-conditioned, set-dependent comparisons required for probabilistic decisions.
Besides, the output head has no class- or set-size-specific parameters, so adding a new option simply adds another row to the computation. This allows an arbitrary number of options in principle, with practical limits imposed by the $K\times K$ attention matrix. 

\subsection{Post-Training}
\label{sec:training}
\noindent\textbf{Unified Decision Supervision.}
For each decision, option $i$ receives a supervision mass $m_i\geq 0$, with positive total mass over the $K$ valid options. Normalizing these masses gives the same target representation $\targetdist=(r_1,\ldots,r_K)^\top\in\Delta^{K-1}$ for hard and soft labels, with $r_i=m_i/\sum_{j=1}^{K}m_j$ for $i=1,\ldots,K$.
A single correct option yields a one-hot target. When multiple options are equally valid, the probability is distributed uniformly among them. For soft labels, the weights come from counts, frequencies, or known probabilities. Padding options are excluded from normalization.

\noindent\textbf{Supervised Fine-Tuning.}
In a batch of $B$ decisions, let $K_n$ denote the number of options in decision $n$, and let $r_{n,i}$ and $p_{n,i}$ denote its target and predicted probabilities for option $i$. We minimize the cross-entropy between the target and predicted distributions.
\begin{equation}
  \mathcal{L}_{\mathrm{SFT}}(\bm{\theta})=
  -\frac{1}{B}\sum_{n=1}^{B}\sum_{i=1}^{K_n}r_{n,i}\log p_{n,i}.
  \label{eq:sft}
\end{equation}
Each decision contributes equally to the batch loss. A one-hot target gives all weight to the correct option, while a soft label distributes supervision in proportion to $r_i$. 
\begin{figure}[t]
  \centering
  \includegraphics[width=\linewidth]{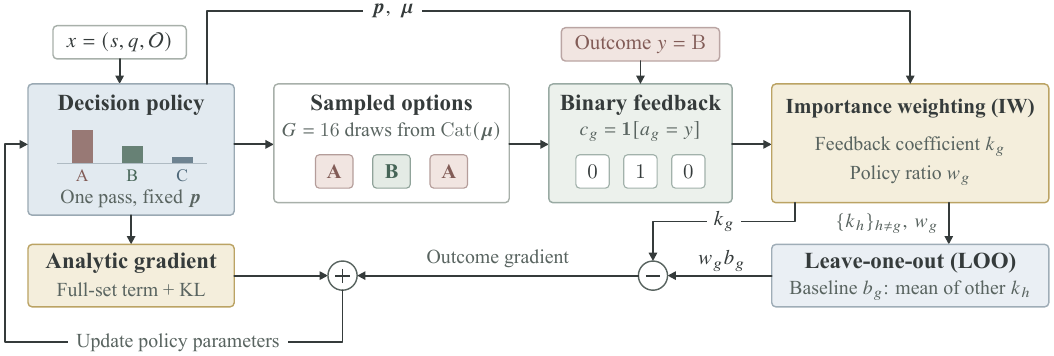}
  \caption{\textbf{Illustration of RLCD.} One forward pass computes the exact gradient term and sampling distribution $\bm{\mu}$ for $G$ draws. Given outcome B, draws A, B, A receive feedback 0, 1, 0. IW and a LOO baseline estimate the remaining gradient in Eq.~\eqref{eq:rl}.}
  \label{fig:rlcd}
\end{figure}

\noindent\textbf{Reinforcement Learning for Calibrated Decisions (RLCD).}
Suppose two options have target probabilities of $0.7$ and $0.3$. The expected match rate $\bm{p}^{\top}\targetdist$ is highest when the model assigns probability one to the first option. To make the predicted probabilities reflect how often each outcome occurs, RLCD combines quadratic and logarithmic losses with weights $\alpha\in[0,1]$ and $1-\alpha$, i.e.,
\begin{equation}
  \ell_\alpha(\bm{p},\targetdist)=\tfrac{\alpha}{2}\|\bm{p}-\targetdist\|_2^2
                   -(1-\alpha)\sum_{i=1}^{K} r_i\log p_i .
  \label{eq:population}
\end{equation}
The quadratic term measures the squared error in each option's probability, and the logarithmic term penalizes assigning very little probability to a likely outcome. 
This combination defines a strictly proper scoring rule whose expected loss is uniquely minimized when the predicted distribution $\bm{p}$ matches the outcome distribution $\targetdist$~\citep{gneiting2007proper}. Any mismatch increases the loss by $\tfrac{\alpha}{2}\|\bm{p}-\targetdist\|_2^2+(1-\alpha)D_{\mathrm{KL}}(\targetdist\|\bm{p})$.

We initialize RLCD from the SFT checkpoint and keep a frozen copy as the reference. Let $\bm{p}_{\mathrm{SFT}}\in\Delta^{K-1}$ denote its prediction for the same decision. With a KL penalty of weight $\beta\geq 0$, we optimize $\mathcal{L}_{\mathrm{RL}}(\bm{\theta})=\ell_\alpha(\bm{p},\targetdist)+\beta D_{\mathrm{KL}}(\bm{p}\|\bm{p}_{\mathrm{SFT}})$. Since a single forward pass produces all $K$ probabilities, we can directly compute the gradients of the prediction-only term $\tfrac{\alpha}{2}\sum_i p_i^2$ and the KL penalty. 
However, for the remaining target-dependent terms, exact soft distributions $\targetdist$ are unavailable in practice, while noisy annotations and single-choice hard labels often introduce imperfect targets.
Thus, rather than deterministically applying $1/p_i$-scaled log-loss gradients to every noisy target, we estimate only these remaining gradient terms from sampled binary feedback, which acts as a soft gradient gate while unifying offline supervision with online feedback.

As shown in Figure~\ref{fig:rlcd}, we construct a sampling distribution $\mu_i=(1-\epsilon)p_i+\epsilon/K$, mixing the current policy with uniform sampling so that every option has a positive sampling probability. We then draw $G$ options $a_g\sim\mathrm{Cat}(\bm{\mu})$ independently with replacement. The feedback source provides a single outcome $y$, equal to the labeled option for a hard target or sampled once from $\targetdist$ for a soft target. Each draw receives binary feedback $c_g=\mathbf{1}[a_g=y]$. All draws share the same outcome and policy forward pass, allowing parallel sampling without additional model evaluations.

Since $\mathbb{E}[c_g\mid a_g=i]=r_i$, we correct for nonuniform sampling via importance weighting, defining the feedback coefficient $k_g=(\alpha p_{a_g}+1-\alpha)c_g/\mu_{a_g}$ and policy ratio $w_g=p_{a_g}/\mu_{a_g}$. For variance reduction, a leave-one-out baseline $b_g=\sum_{h\ne g}k_h/(G-1)$ gives the adjusted coefficient $k_g-w_gb_g$. Combining this term with the exact gradient yields
\begin{equation}
\begin{aligned}
 \widehat{\nabla}_{\bm{\theta}}\mathcal{L}_{\mathrm{RL}}={}&
   \nabla_{\bm{\theta}}\!\left[\tfrac{\alpha}{2}\|\bm{p}\|_2^2+
              \beta D_{\mathrm{KL}}(\bm{p}\|\bm{p}_{\mathrm{SFT}})\right]\\
 &-\tfrac{1}{G}\sum_{g=1}^{G}(k_g-w_g b_g)
                       \nabla_{\bm{\theta}}\log p_{a_g}.
\end{aligned}
\label{eq:rl}
\end{equation}
For a fixed outcome $y$, the baseline $b_g$ is independent of $a_g$, so its gradient contribution has zero expectation. Thus, the resulting estimator is unbiased for $\nabla_{\bm{\theta}}\mathcal{L}_{\mathrm{RL}}$ when averaged over outcomes and sampled options. We provide the derivation in Appendix~\ref{app:derivations}. Algorithm~\ref{alg:rlcd} shows the RLCD update from binary outcome feedback.
\begin{algorithm}[tb]
\caption{One RLCD update from binary outcome feedback}
\label{alg:rlcd}
\small
\renewcommand{\algorithmicrequire}{\textbf{Input}}
\begin{algorithmic}[1]
\REQUIRE Decisions $\{x_n\}_{n=1}^{B}$ (new or replayed), actor $\decisionmodel$,
    frozen reference $f_{\mathrm{SFT}}$, feedback source, and $(\alpha,\beta,\epsilon,G)$.
\FOR{$n=1,\ldots,B$}
    \STATE $\bm p\gets\decisionmodel(x_n)$,\quad
           $\bm p_{\mathrm{SFT}}\gets f_{\mathrm{SFT}}(x_n)$
           \hfill\textit{Score all $K_n$ options, Eq.~\eqref{eq:readout}.}
    \STATE $\bm g_{\mathrm{exact}}\gets\text{first term of Eq.~\eqref{eq:rl}}$.
           \hfill\textit{Exact quadratic and KL gradients.}
    \STATE $\bm\mu\gets(1-\epsilon)\stopgrad(\bm p)+(\epsilon/K_n)\bm 1$.
    \STATE $a_1,\ldots,a_G\overset{\mathrm{iid}}{\sim}\mathrm{Cat}(\bm\mu)$.
    \STATE Feedback source draws $y\sim\mathrm{Cat}(\targetdist)$ once and returns
           $c_g=\mathbf 1[a_g=y]$ for $g=1,\ldots,G$.
    \STATE $w_g\gets p_{a_g}/\mu_{a_g}$,\quad
           $k_g\gets c_g\big[\alpha w_g+(1-\alpha)/\mu_{a_g}\big]$.
           \hfill\textit{Importance weighting.}
    \STATE $b_g\gets\operatorname{mean}_{h\ne g} k_h$ if $G>1$, else $0$.
           \hfill\textit{Leave-one-out baseline.}
    \STATE $\bm g_{\mathrm{sample}}\gets\text{signed second term of Eq.~\eqref{eq:rl}}$.
           \hfill\textit{Hold $k_g-w_g b_g$ fixed.}
    \STATE $\widehat{\bm g}_n\gets\bm g_{\mathrm{exact}}+\bm g_{\mathrm{sample}}$.
\ENDFOR
\STATE Update $\bm\theta$ with AdamW using $B^{-1}\sum_{n=1}^{B}\widehat{\bm g}_n$.
\end{algorithmic}
\end{algorithm}

\subsection{Data Curation and Allocation}
\label{sec:data}
We combine text decisions from the public OpenJevData-140K dataset~\citep{openjevdata2026} with multimodal training data we construct. 
Our sources include public visual question answering datasets, human judgments of response quality, robot and game trajectories, and outcomes we collect by executing actions in simulators. 
In each visual decision, the state contains an image or a sequence of keyframes in temporal order, and the query specifies the task goal for the supplied options.
For visual question answering and response comparisons, we use the original answer and preference annotations, including judgments of which response better describes a scene.
For interactive tasks, we turn game and robot trajectories into next-action decisions, using only observations recorded before the action and taking the executed action as the target.
To compare the effects of different actions, we execute them from the same simulator state and derive supervision from the resulting task outcomes.

To reduce shortcuts and label noise, we remove text-solvable or similarity-matched visual examples, discard conflicting duplicates, and validate targets against evidence, option identities, and masks.
To prevent leakage, we group related renderings, shared images, and episode windows before splitting, screen for evaluation overlaps, and exclude all evaluation games and embodied environments.
Most decisions contain two to eight options, with source- and episode-level quotas preserving coverage of less frequent tasks. Appendix~\ref{app:data} provides further details on the data sources and filtering procedure.

Our additional experiments show that SFT is more sample-efficient with hard labels, whereas RLCD benefits more from soft labels (Appendix~\ref{app:additional}).
Thus, we allocate the data accordingly, i.e., about 80\% of the 116K SFT examples have hard labels, with the visual subset mainly consisting of image questions and pairwise response preferences. In contrast, only about 30\% of the 126K RLCD examples have hard labels.
We prioritize robot and game decisions for RLCD when the observations support multiple plausible actions but do not clearly distinguish their relative quality.

\subsection{Implementation and Infrastructure}
\label{sec:infrastructure}
In both SFT and RLCD, we update the language backbone and decision head while keeping the vision backbone frozen~\citep{liu2023llava,meng2025mmeureka,shen2025sophia}.
For both sizes, we train for one epoch per stage using AdamW with a global batch size of 1,024 decisions.
In SFT, the learning rates are $4\times10^{-6}$ for the language backbone and $4\times10^{-5}$ for the decision head. In RLCD, we lower them to $7.5\times10^{-7}$ and $7.5\times10^{-6}$, respectively.
We use $G=16$ draws per decision, a loss mixing weight of $\alpha=0.5$, a uniform sampling probability of $\epsilon=0.1$, and a KL penalty $\beta=0.01$.
Appendix~\ref{app:derivations} gives the detailed configurations of both models.

Because the decision head scores options jointly, we keep all options of a decision on the same rank while sharding model parameters across devices.
Option branches of similar length are encoded in parallel, and their final hidden states are then regrouped by decision for joint scoring.
The loss is averaged over decisions to ensure that decisions with more options do not receive greater weight.

At inference, we prefill the shared state and query once and evaluate all $K$ option suffixes in parallel from the cached prefix, which for Qwen3.5 stores attention KV, linear-attention convolutional and recurrent states, and the position offset.
For prefix length $L$ and suffix lengths $C_1,\ldots,C_K$, prefix reuse reduces backbone tokens from $KL+\sum_{i=1}^{K}C_i$ to $L+\sum_{i=1}^{K}C_i$, passing only the $K$ terminal states to the decision head.

\begin{table*}[t]
\reporttable
\caption{\textbf{Overall performance on interactive and visual benchmarks.} Task scores are percentages. \ranklegend{} ``--'' denotes not applicable. $^{\dagger}$~Results from the Decision Index skill score (Appendix~\ref{app:protocol}). $^{*}$~Autoregressive answer generation with reasoning disabled. $^{\ddagger}$~Next-token probabilities normalized over the answer letters.}
\label{tab:vision}
\scalebox{1.0}{%
\begin{tabularx}{\linewidth}{lcccYYcYY}
\toprule
\multirow{2}{*}[-0.5ex]{\tablehead{Model}} & \multicolumn{1}{c}{\tablehead{P95 Latency}} & \multicolumn{2}{c}{\tablehead{GameWorld}} & \multicolumn{2}{c}{\tablehead{EmbodiedBench}} & \multicolumn{1}{c}{\tablehead{VLRewardBench}} & \multicolumn{2}{c}{\tablehead{Visual Knowledge}} \\
\cmidrule(lr){2-2}\cmidrule(lr){3-4}\cmidrule(lr){5-6}\cmidrule(lr){7-7}\cmidrule(lr){8-9}
 & \tablehead{\mbox{(s)$\downarrow$}} & \tablehead{\mbox{PG$\uparrow$}} & \tablehead{\mbox{SR$\uparrow$}} & \tablehead{\mbox{ALFRED$\uparrow$}} & \tablehead{\mbox{Habitat$\uparrow$}} & \tablehead{\mbox{Macro Acc.$\uparrow$}} & \tablehead{\mbox{MMMU-Pro$\uparrow$}} & \tablehead{\mbox{InfoVQA$\uparrow$}} \\
\midrule
\rowcolor{tablegroupbg}\multicolumn{9}{c}{\tablehead{Proprietary Models}} \\
Jev 1.14 & -- & 30.51 & 11.18 & 16.67 & 21.33 & 75.41 & \bestscore{54.74} & 94.67 \\
Perplexity Decider-V1 & -- & 30.06 & 8.24 & 26.00 & 26.67 & \secondscore{75.49} & 33.93 & 81.56 \\
Clef-Omni & -- & 22.13 & 5.29 & 6.33 & 28.00 & 62.71 & 32.31\rlap{$^\dagger$} & 90.57\rlap{$^\dagger$} \\
GPT-6 Luna Decisions & -- & \secondscore{36.03} & \secondscore{17.06} & 19.33 & 32.67 & 63.30 & 37.17 & 84.02 \\
GPT-5.5$^{*}$ & -- & 34.00 & 15.88 & \secondscore{42.00} & \secondscore{47.33} & 69.09 & 52.49 & 91.39 \\
GPT-4o$^{*}$ & -- & 30.15 & 11.76 & 27.00 & 30.33 & 62.01 & 38.03 & 88.11 \\
\midrule
\rowcolor{tablegroupbg}\multicolumn{9}{c}{\tablehead{Open-source Backbone (4B)}} \\
Qwen3.5-4B$^{\ddagger}$ & \secondscore{0.33} & 16.32 & 5.88 & 0.67 & 5.67 & 61.10 & 25.49 & 82.79 \\
Qwen3.5-4B$^{*}$ & 60.49 & 9.02 & 6.47 & 1.33 & 15.33 & 12.27 & 27.80 & 70.90 \\
Qwen3-VL-4B-Instruct$^{\ddagger}$ & \secondscore{0.33} & 24.03 & 8.82 & 3.33 & 13.67 & 63.67 & 30.87 & 82.38 \\
Qwen3-VL-4B-Instruct$^{*}$ & 0.44 & 24.30 & 9.41 & 3.67 & 13.67 & 46.23 & 30.52 & 81.56 \\
Intern-Decision-4B & 0.37 & 23.41 & 7.65 & 13.33 & 10.00 & 68.55 & 38.55\rlap{$^\dagger$} & 87.30\rlap{$^\dagger$} \\
H2O-Lightning-4B & 1.00 & 27.36 & 9.41 & 9.67 & 19.00 & 68.12 & 37.86 & 91.80 \\
\rowcolor{tableoursbg}\textbf{LightDecision-1 4B} & \bestscore{0.31} & 31.18 & 10.59 & 13.33 & 38.67 & 68.51 & 38.67 & 92.21 \\
\midrule
\rowcolor{tablegroupbg}\multicolumn{9}{c}{\tablehead{Open-source Backbone (27B)}} \\
Qwen3.6-27B$^{\ddagger}$ & 0.79 & 23.54 & 8.82 & 27.33 & 26.67 & 72.00 & 44.22 & 91.39 \\
Qwen3.6-27B$^{*}$ & 188.17 & 26.97 & 11.76 & 30.33 & 32.33 & 36.20 & 18.61 & 84.84 \\
Qwen3.5-27B$^{\ddagger}$ & 0.79 & 24.62 & 6.47 & 21.33 & 20.67 & 72.67 & 45.90 & 92.21 \\
Qwen3.5-27B$^{*}$ & 144.26 & 31.79 & 12.94 & 22.33 & 26.33 & 42.17 & 36.42 & 87.30 \\
WeMM-JEV-27B & 1.08 & 33.91 & 12.94 & 37.67 & 34.00 & \bestscore{75.77} & 52.54 & \secondscore{95.08} \\
\rowcolor{tableoursbg}\textbf{LightDecision-1 27B} & 0.76 & \bestscore{41.33} & \bestscore{17.65} & \bestscore{43.00} & \bestscore{48.67} & 73.21 & \secondscore{53.18} & \bestscore{95.90} \\
\bottomrule
\end{tabularx}%
}
\end{table*}

\section{Experiments}
\label{sec:evaluation}
\subsection{Experimental Setup}
\noindent\textbf{Baselines.}
We compare \methodname{} with decision models and general-purpose vision--language models.
The decision baselines include the proprietary models Jev~1.14~\citep{typesafe2026jev}, Perplexity Decider-V1~\citep{perplexity2026decider}, Clef-Omni~\citep{cloudflare2026clef} and GPT-6 Luna Decisions~\citep{openai2026decisions}.
The open-weight decision baselines are Intern-Decision-4B~\citep{zhu2026intern}, H2O-Lightning-4B~\citep{h2o2026lightning}, and WeMM-JEV-27B~\citep{wemmjevteam2026wemm}.
The general baselines are GPT-5.5~\citep{openai2026gpt55}, GPT-4o~\citep{openai2024gpt4o}, Qwen3.6-27B~\citep{qwen2026qwen36}, Qwen3.5-4B~\citep{qwen2026qwen35}, Qwen3.5-27B~\citep{qwen2026qwen35}, and Qwen3-VL-4B-Instruct~\citep{qwen2025qwen3vl4b}, which generate their answers autoregressively ($^{*}$). For the four Qwen models, we also normalize next-token probabilities over the answer letters to obtain an option distribution ($^{\ddagger}$).
We evaluate all general models without a reasoning mode because decision models generate a probability distribution over options only in one forward pass. Appendix~\ref{app:baselines} gives the model settings.

\noindent\textbf{Evaluation.}
We evaluate interactive tasks, visual question answering, and reward modeling.
On GameWorld~\citep{gameworld}, we report progress (PG) and success rate (SR). PG measures the average fraction of progress toward the goal, based on the best score achieved and clipped to $[0,1]$. SR is the fraction of successfully completed tasks.
On EmbodiedBench~\citep{embodiedbench}, we report SR separately for ALFRED and Habitat subsets.
MMMU-Pro (vision) and InfographicVQA measure answer accuracy~\citep{mmmupro,infographicvqa,decisionindex}, and VL-RewardBench~\citep{vlrewardbench} reports macro accuracy across general, hallucination, and reasoning domains.
Higher scores are better. For decision models, the number of options is not fixed in advance. For \methodname{}, option names are out of distribution. Appendix~\ref{app:protocol} provides more details.

\noindent\textbf{Latency.} 
We measure P95 latency on MMMU-Pro with batch size $1$ on a single GPU (\textasciitilde$1$~PFLOPS peak BF16), encoding the image and prefilling the prefix once per question for fairness.
Decision baselines and logit-scoring VLMs score packed options in one forward pass, autoregressive VLMs decode sequentially, and \methodname{} evaluates option branches in parallel from the shared prefix.
Synchronized device timing includes vision encoding, prefilling, and all forward passes, excluding data preprocessing, host-to-device transfer, and answer parsing.

\subsection{Main Results}
\label{sec:visual-results}
\begin{figure*}[t]
  \centering
  \includegraphics[width=\textwidth]{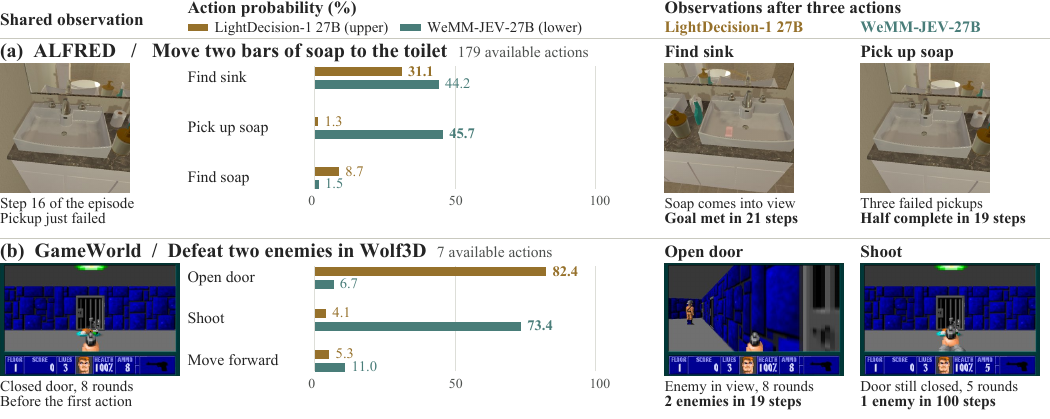}
  \caption{\textbf{Case study comparing two models.} At their first differing action, both models share the same observation, query, options and history. Bars show probabilities for their top-3 actions. Bold values mark their choices. Right panels show the scene three actions later.}
  \label{fig:interaction}
\end{figure*}

\textbf{Comparison with Decision Models.} 
As shown in Table~\ref{tab:vision}, \methodname{} 27B consistently outperforms both open-weight and proprietary decision baselines across GameWorld and EmbodiedBench, and achieves highly competitive results on VL-RewardBench and visual knowledge tasks.
Meanwhile, \methodname{} 4B achieves the strongest overall performance among models of comparable scale, and even surpasses larger 27B or proprietary decision models in interactive environments.
Although some existing decision baselines perform well on tasks with few static options such as VL-RewardBench, they degrade sharply in interactive environments where action selection from hundreds of options must adapt to the current state and feedback. For example, Figure~\ref{fig:interaction} shows that WeMM-JEV-27B struggles to adapt its actions to feedback and the current scene. In ALFRED, it assigns 45.7\% probability to repeating a pickup that just failed, while \methodname{} switches to ``Find sink''. In Wolf3D, it favors shooting at a closed door (73.4\%), while \methodname{} correctly chooses ``Open door''. These choices lead to markedly different outcomes, with \methodname{} completing the two tasks in 21 and 19 steps, respectively.
Thus, \methodname{} establishes a strong advantage over existing decision models, bridging the gap between static visual understanding and sequential interactive control across model scales.

\textbf{Comparison with General Models.}
Compared with normalizing next-token probabilities over answer letters on the corresponding Qwen3.5 backbones ($\ddagger$), \methodname{} achieves substantial gains across all benchmarks at both scales, indicating that reusing pretrained vocabulary tokens for probability prediction disrupts visual-semantic representations and introduces token biases.
The same advantage holds against autoregressive generation ($*$). Both the 4B and 27B models outperform their corresponding autoregressive Qwen baselines across all metrics, and even compared with concise instruction-tuned generators and frontier proprietary models, \methodname{} 27B surpasses GPT-5.5 and GPT-4o on all seven reported metrics.
Therefore, projecting hidden states into a learned decision space and training with our recipe strengthens decision quality over general models without requiring reasoning traces.

\begin{figure}[t]
  \centering
  \includegraphics[width=\linewidth]{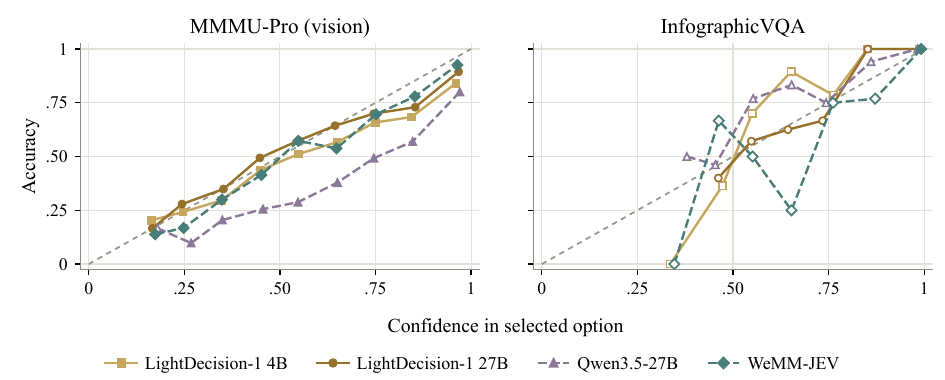}
  \caption{\textbf{Probability calibration on visual questions.} Each point compares mean confidence with accuracy for one bin. The gray diagonal marks perfect calibration. Points below it indicate overconfidence and points above it indicate underconfidence. 
  Hollow markers denote bins with fewer than 20 questions.}
  \label{fig:visual-calibration}
\end{figure}

\textbf{Inference Latency.} 
Low inference latency makes System-1 decision models well suited for real-time interaction. Almost all single-pass models achieve much lower P95 latency than autoregressive baselines.
Among all models, LightDecision-1 4B achieves the lowest latency, while LightDecision-1 27B is the fastest among 27B models. 
Notably, \methodname{} achieves even lower P95 latency than the backbone scoring, because caching the shared prefix avoids cross-option attention in the deep backbone and evaluates short option branches in parallel.
In conclusion, \methodname{} combines the lowest decision latency with the strongest task performance at each scale.

\subsection{Architecture Analysis}
\label{sec:analysis}

\noindent\textbf{Probability Calibration.}
A useful decision model should make accurate choices and indicate how much to trust them. When it assigns 80\% confidence to a choice, it should be right about 80\% of the time. Figure~\ref{fig:visual-calibration} compares the probability assigned to the selected option with how often that option is correct.
On MMMU-Pro, \methodname{} 27B follows the diagonal much more closely than Qwen3.5-27B, which is consistently overconfident across most confidence levels. On InfographicVQA, our model also shows relatively stable calibration, while WeMM-JEV exhibits larger fluctuations across confidence bins. Overall, \methodname{} provides more reliable confidence estimates across both visual tasks, making its predictions more useful for downstream decisions.

\noindent\textbf{Permutation Equivariance.}
Reliable decisions should depend solely on option semantics rather than arbitrary presentation order.
Across cyclic option rotations on 48 InfographicVQA questions, Table~\ref{tab:visual-order} shows that all baseline decision models exhibit severe order sensitivity, with probability drifts up to $0.6311$ and selection flips on up to $12/48$ questions.
In contrast, both scales of \methodname{} maintain zero selection flips, while the negligible drift stems solely from floating-point rounding in computations.

\begin{figure}[t]
  \centering
  \captionsetup{skip=4pt}
  \includegraphics[width=\linewidth,trim=0 10bp 0 0,clip]{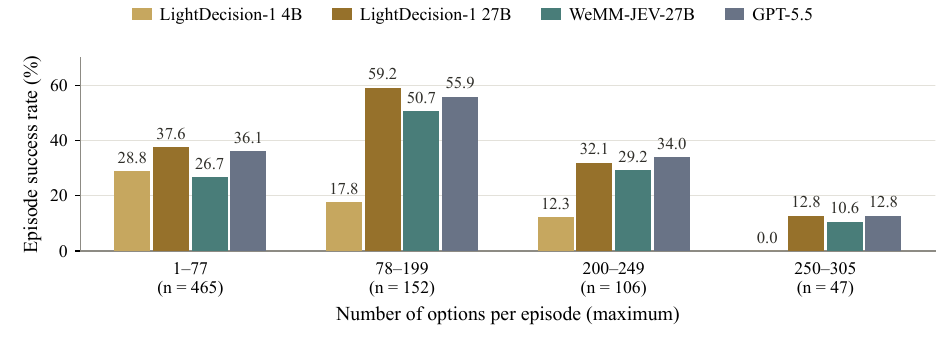}
  \caption{\textbf{Interactive task success by option count.} Bars show success rates pooled across GameWorld, ALFRED, and Habitat. Each task counts once, with the same bin for every model. GameWorld uses the largest option set in its reference trace.}
  \label{fig:action-set-success}
\end{figure}

\begin{table}[t]
\reporttable
\caption{\textbf{Inference latency with and without prefix sharing.} Expanded execution encodes the prefix for each option. Speedup is the ratio of expanded to shared median latency over three runs.}
\label{tab:time}
\small
\setlength{\tabcolsep}{3pt}
\renewcommand{\arraystretch}{1.12}
\begin{tabularx}{\linewidth}{l c *{6}{Y}}
\toprule
\multirow{2}{*}{\tablehead{Shared state}} & \multirow{2}{*}{\tablehead{Options}} &
  \multicolumn{3}{c}{\tablehead{LightDecision-1 4B}} &
  \multicolumn{3}{c}{\tablehead{LightDecision-1 27B}} \\
\cmidrule(lr){3-5}\cmidrule(lr){6-8}
 & & \tablehead{Expanded (s)} & \tablehead{Shared (s)} & \tablehead{Speedup}
   & \tablehead{Expanded (s)} & \tablehead{Shared (s)} & \tablehead{Speedup} \\
\midrule
\multirow{2}{*}{Image + 128 tokens} & 16 & 0.294 & 0.247 & 1.19$\times$ & 0.936 & 0.446 & 2.10$\times$ \\
 & 128 & 2.195 & 1.371 & 1.60$\times$ & 7.429 & 2.752 & 2.70$\times$ \\
\addlinespace[2pt]
\multirow{2}{*}{2,048 tokens} & 16 & 0.859 & 0.234 & 3.67$\times$ & 3.920 & 0.578 & 6.78$\times$ \\
 & 128 & 6.851 & 1.458 & 4.70$\times$ & 31.529 & 2.858 & 11.03$\times$ \\
\bottomrule
\end{tabularx}
\end{table}

\begin{table}[t]
\reporttable
\caption{\textbf{Permutation equivariance analysis.} Worst drift is the largest change in any option's probability after matching by identity. A flip changes the selected option relative to the original order. The last column counts questions with at least one flip.}
\label{tab:visual-order}
\small
\setlength{\tabcolsep}{5pt}
\renewcommand{\arraystretch}{1.12}
\begin{tabularx}{\linewidth}{X >{\centering\arraybackslash}p{.17\linewidth} *{2}{>{\centering\arraybackslash}p{.235\linewidth}}}
\toprule
\tablehead{Model} & \tablehead{Worst drift $\downarrow$} & \tablehead{Selection flip rate $\downarrow$} & \tablehead{Questions with flips $\downarrow$} \\
\midrule
WeMM-JEV-27B & 0.2631 & 0.69\% (1/144) & 1/48 \\
H2O-Lightning-4B & 0.2797 & 8.33\% (12/144) & 6/48 \\
Intern-Decision-4B & 0.6311 & 15.28\% (22/144) & 12/48 \\
\midrule
\rowcolor{tableoursbg}\textbf{LightDecision-1 4B} & $1.79\times10^{-7}$ & 0\% (0/144) & 0/48 \\
\rowcolor{tableoursbg}\textbf{LightDecision-1 27B} & $1.19\times10^{-7}$ & 0\% (0/144) & 0/48 \\
\bottomrule
\end{tabularx}
\end{table}

\begin{table}[!htbp]
\reporttable
\caption{\textbf{Ablations on LightDecision-1 4B.} The three groups isolate cross-option attention, the second-stage training objective, and the RLCD gradient estimator. The final row reports the full model.}
\label{tab:ablations}
\small
\setlength{\tabcolsep}{5pt}
\renewcommand{\arraystretch}{1.12}
\begin{tabularx}{\linewidth}{X *{3}{>{\centering\arraybackslash}p{.161\linewidth}}}
\toprule
\multirow{2}{*}[-0.5ex]{\tablehead{Variant}} & \multicolumn{2}{c}{\tablehead{GameWorld}} & \tablehead{InfoVQA} \\
\cmidrule(lr){2-3}\cmidrule(lr){4-4}
 & \tablehead{PG $\uparrow$} & \tablehead{SR $\uparrow$} & \tablehead{Acc. $\uparrow$} \\
\midrule
\rowcolor{tablegroupbg}\multicolumn{4}{c}{\tablehead{Joint option modeling}} \\
Independent bilinear readout & 22.87 & 7.06 & 87.70 \\
Self-only attention & 29.85 & 10.00 & 91.39 \\
\rowcolor{tablegroupbg}\multicolumn{4}{c}{\tablehead{Post-training strategy}} \\
SFT only & 19.42 & 6.47 & 86.48 \\
SFT + continued cross-entropy & 20.48 & 7.06 & 90.16 \\
SFT + full-target quadratic--log loss & 25.82 & 8.24 & 90.57 \\
\rowcolor{tablegroupbg}\multicolumn{4}{c}{\tablehead{Effectiveness of RLCD}} \\
Full sampling & 29.24 & 9.41 & 90.98 \\
PPO with ratio clipping & 27.15 & 8.82 & 90.16 \\
\midrule
\rowcolor{tableoursbg}\textbf{LightDecision-1 4B (full)} & 31.18 & 10.59 & 92.21 \\
\bottomrule
\end{tabularx}
\end{table}

\noindent\textbf{Generalization to Larger Sets.}
Both training stages use at most 77 options per decision, yet \methodname{} 27B generalizes to episodes with up to 305 available actions. Figure~\ref{fig:action-set-success} shows it outperforms WeMM-JEV-27B across all option-count ranges and remains competitive with GPT-5.5, even on the largest action sets. Its advantage over the 4B model on larger sets also shows that backbone scaling improves the ability to compare many open-vocabulary options. These results show the effectiveness of our approach in generalizing to action sets far beyond the training range.

\noindent\textbf{Parallel Acceleration.}
Table~\ref{tab:time} evaluates prefix sharing against per-option prefix encoding under fixed weights, inputs, and microbatch size. Aside from minor caching overhead on short text-only prompts, sharing the prefix substantially accelerates multimodal and long-context decisions.
With 128 options, increasing the text state from 128 to 2,048 tokens raises the 27B speedup from $1.10\times$ to $11.03\times$. With the shorter state, doubling the option count from 512 to 1,024 yields little further change at either scale.

\subsection{Ablation Study}
\label{sec:ablation}
Table~\ref{tab:ablations} ablates three core design choices on \methodname{}-4B across GameWorld and InfoVQA, i.e., joint option modeling, post-training strategy, and the RLCD estimator.

\noindent\textbf{Joint Option Modeling.}
Under identical training data and hyperparameters, we compare the full model against \emph{independent bilinear readout}, which scores projected option vectors prior to the set-attention blocks, and \emph{self-only attention}, which retains both blocks and feed-forward layers while masking attention to the diagonal.
\emph{Independent bilinear readout} degrades sharply without decision-head refinement, while \emph{self-only attention} recovers most of the gap yet slightly underperforms the full model on interdependent options.
Thus, cross-option interaction is essential.

\noindent\textbf{Post-training Strategy.}
Starting from the \emph{SFT only} policy, we compare second-stage training under identical data, optimizer settings, and KL regularization across \emph{continued cross-entropy}, \emph{full-target quadratic--log loss} (exact gradient of Eq.~\eqref{eq:population}), and RLCD.
With minimal embodied data, \emph{SFT only} lags on GameWorld, and \emph{continued cross-entropy} boosts InfoVQA but barely helps GameWorld due to non-exploratory imitation.
\textit{Full-target quadratic--log loss} improves over cross-entropy, yet still falls short of RLCD. Deterministic full-vector differentiation applies steep $1/p_i$ log-loss gradients to every low-probability noisy annotation or single-choice hard label, causing overfitting. In contrast, the sampled binary feedback acts as a soft gradient gate that filters outlier targets while preserving unbiased calibration in expectation.
Thus, learning from sampled binary feedback in RLCD mitigates target noise and promotes exploration.

\noindent\textbf{Effectiveness of RLCD.}
Under identical settings from the same SFT checkpoint, we compare RLCD with \emph{full sampling}, which samples both gradient terms, and \emph{PPO with ratio clipping}, which reuses sampled feedback across updates.
Although \emph{full sampling} is unbiased, sampling the prediction-only term adds gradient variance and weakens performance. \emph{PPO with ratio clipping} performs worse because clipping introduces optimization bias and multi-step ratio reuse slows updates.
In contrast, RLCD computes the prediction-only and KL gradients analytically and applies a leave-one-out baseline only to the outcome term, achieving cleaner optimization and the strongest results.

\FloatBarrier
\section{Conclusion}
\label{sec:conclusion}
This paper presents \methodname{}, a family of 4B and 27B open-vocabulary probabilistic multimodal System-1 decision models in a single forward pass.
By caching the shared state and query prefix across parallel option branches and comparing candidates through a permutation-equivariant set-attention head, \methodname{} avoids vocabulary-token biases and eliminates order sensitivity.
Through our two-stage SFT--RLCD recipe, it further combines exact analytic gradients with unbiased leave-one-out estimates from parallel binary feedback under a strictly proper scoring rule to learn calibrated probabilities from both hard and soft supervision.
Extensive experiments show that \methodname{} leads open-weight models and surpasses frontier proprietary systems on several benchmarks, delivering better calibration, generalization beyond 300 options, and sub-second latency, while ablations confirm the benefits of joint option interaction, prefix reuse, and our variance-reduced RLCD estimator.
Despite these advances, \methodname{} still faces challenges in longer-horizon and visually complex environments, where current training coverage remains limited. In future work, we plan to scale training to richer, more complex scenarios and enable the model to explore and learn directly through interaction.

\phantomsection
\addcontentsline{toc}{section}{Contributions}
\section*{Contributions}

Names within each role are listed in reverse alphabetical order by surname (Z to A).

\begingroup
\renewcommand{\arraystretch}{1.25}
\noindent\begin{tabularx}{\linewidth}{@{}>{\bfseries}l@{\hspace{1.5em}}>{\raggedright\arraybackslash}X@{}}
  Project Leader & Junhao~Shen\textsuperscript{1} \\
  Contributors &
    Zishan~Xu\textsuperscript{5}, Xinye~Li\textsuperscript{1},
    Jinyang~Huang\textsuperscript{4},\newline
    Mingqi~Gao\textsuperscript{2},
    Enjun~Du\textsuperscript{3}, Zirong~Chen\textsuperscript{2} \\
  Corresponding Authors &
    Zhanqiu~Zhang\textsuperscript{6}, Yiwen~Guo\textsuperscript{7} \\
  Academic Advisor & Hong~Cheng\textsuperscript{1}
\end{tabularx}
\endgroup

{\footnotesize
  \textsuperscript{1}The Chinese University of Hong Kong \quad
  \textsuperscript{2}Tsinghua University \quad
  \textsuperscript{3}The University of Hong Kong\\
  \textsuperscript{4}Central South University \quad
  \textsuperscript{5}Shanghai Jiao Tong University \quad
  \textsuperscript{6}LIGHTSPEED \quad
  \textsuperscript{7}Independent Researcher\par}

\textbf{Contact}\quad \email{shen.junhao@outlook.com}

\clearpage
\phantomsection
\addcontentsline{toc}{section}{\refname}
\begingroup
  \setlength{\bibsep}{2pt}
  \bibliographystyle{lightspeed_techreport}
  \bibliography{references}

\begin{thebibliography}{76}
\providecommand{\natexlab}[1]{#1}
\providecommand{\url}[1]{\texttt{#1}}
\expandafter\ifx\csname urlstyle\endcsname\relax
  \providecommand{\doi}[1]{doi: #1}\else
  \providecommand{\doi}{doi: \begingroup \urlstyle{rm}\Url}\fi

\bibitem[Almeida(2026)]{typesafe2026jev}
Diogo Almeida.
\newblock Introducing system one models \& {Jev}, 2026.
\newblock URL \url{https://typesafe.ai/blog/introducing-system-one-models-and-jev}.

\bibitem[Chen et~al.(2024{\natexlab{a}})Chen, Qin, Zhang, Chen, Xu, and Che]{chen2024m3cot}
Qiguang Chen, Libo Qin, Jin Zhang, Zhi Chen, Xiao Xu, and Wanxiang Che.
\newblock {M}$^3${C}o{T}: A novel benchmark for multi-domain multi-step multi-modal chain-of-thought.
\newblock In \emph{Proceedings of the 62nd Annual Meeting of the {A}ssociation for {C}omputational {L}inguistics (Volume 1: Long Papers)}, pp.\  8199--8221, Bangkok, Thailand, 2024{\natexlab{a}}.

\bibitem[Chen et~al.(2026{\natexlab{a}})Chen, Ge, Ge, Ding, Li, Wang, Xu, Shan, and Liu]{chen2026egoplan}
Yi~Chen, Yuying Ge, Yixiao Ge, Mingyu Ding, Bohao Li, Rui Wang, Ruifeng Xu, Ying Shan, and Xihui Liu.
\newblock {EgoPlan-Bench}: Benchmarking multimodal large language models for human-level planning.
\newblock \emph{International Journal of Computer Vision}, 134\penalty0 (3):\penalty0 118, 2026{\natexlab{a}}.

\bibitem[Chen et~al.(2024{\natexlab{b}})Chen, Wu, Wang, Su, Chen, Xing, Zhong, Zhang, Zhu, Lu, Li, Luo, Lu, Qiao, and Dai]{chen2024internvl}
Zhe Chen, Jiannan Wu, Wenhai Wang, Weijie Su, Guo Chen, Sen Xing, Muyan Zhong, Qinglong Zhang, Xizhou Zhu, Lewei Lu, Bin Li, Ping Luo, Tong Lu, Yu~Qiao, and Jifeng Dai.
\newblock {InternVL}: Scaling up vision foundation models and aligning for generic visual-linguistic tasks.
\newblock In \emph{Proceedings of the IEEE/CVF Conference on Computer Vision and Pattern Recognition}, pp.\  24185--24198, Seattle, WA, 2024{\natexlab{b}}.

\bibitem[Chen et~al.(2026{\natexlab{b}})Chen, Ye, Zhang, Du, Pu, Wang, Zuo, Duan, Ma, and Zhang]{chen2026snapbench}
Zirong Chen, Fuda Ye, Kuan Zhang, Enjun Du, Junfu Pu, Xinlei Wang, Xinyu Zuo, Lisheng Duan, Jin Ma, and Yongqi Zhang.
\newblock {SnapBench}: Benchmarking snap-and-ask multimodal retrieval for mobile interactions.
\newblock In \emph{Findings of the Association for Computational Linguistics: EMNLP 2026}, pp.\  20143--20178, Budapest, Hungary, 2026{\natexlab{b}}.

\bibitem[Cheng et~al.(2024)Cheng, Sun, Chu, Xu, Li, Zhang, and Wu]{cheng2024seeclick}
Kanzhi Cheng, Qiushi Sun, Yougang Chu, Fangzhi Xu, Yantao Li, Jianbing Zhang, and Zhiyong Wu.
\newblock {SeeClick}: Harnessing {GUI} grounding for advanced visual {GUI} agents.
\newblock In \emph{Proceedings of the 62nd Annual Meeting of the {A}ssociation for {C}omputational {L}inguistics (Volume 1: Long Papers)}, pp.\  9313--9332, Bangkok, Thailand, 2024.

\bibitem[Chevalier-Boisvert et~al.(2023)Chevalier-Boisvert, Dai, Towers, Perez-Vicente, Willems, Lahlou, Pal, Castro, and Terry]{chevalierboisvert2023minigrid}
Maxime Chevalier-Boisvert, Bolun Dai, Mark Towers, Rodrigo Perez-Vicente, Lucas Willems, Salem Lahlou, Suman Pal, Pablo~Samuel Castro, and Jordan Terry.
\newblock {Minigrid} \& {Miniworld}: Modular \& customizable reinforcement learning environments for goal-oriented tasks.
\newblock In \emph{Advances in Neural Information Processing Systems 36}, pp.\  73383--73394, New Orleans, LA, 2023.

\bibitem[{Cloudflare}(2026)]{cloudflare2026clef}
{Cloudflare}.
\newblock Introducing {Clef-omni} with full multimodality, plus a faster {Clef} and a cheaper {Clef-flash}, 2026.
\newblock URL \url{https://blog.cloudflare.com/clef-faster-cheaper-multimodal/}.

\bibitem[Cobbe et~al.(2020)Cobbe, Hesse, Hilton, and Schulman]{cobbe2020procgen}
Karl Cobbe, Chris Hesse, Jacob Hilton, and John Schulman.
\newblock Leveraging procedural generation to benchmark reinforcement learning.
\newblock In \emph{Proceedings of the 37th International Conference on Machine Learning}, pp.\  2048--2056, Virtual, 2020.

\bibitem[Dai et~al.(2023)Dai, Li, Li, Tiong, Zhao, Wang, Li, Fung, and Hoi]{dai2023instructblip}
Wenliang Dai, Junnan Li, Dongxu Li, Anthony Meng~Huat Tiong, Junqi Zhao, Weisheng Wang, Boyang Li, Pascale Fung, and Steven Hoi.
\newblock {InstructBLIP}: Towards general-purpose vision-language models with instruction tuning.
\newblock In \emph{Advances in Neural Information Processing Systems 36}, pp.\  49250--49267, New Orleans, LA, 2023.

\bibitem[{Decision Index Contributors}(2026)]{decisionindex}
{Decision Index Contributors}.
\newblock {Decision Index}.
\newblock \url{https://github.com/apolinario/decision-index/tree/25c6ec7b6b5f}, 2026.

\bibitem[Deu{\ss}er et~al.(2026)Deu{\ss}er, Sparrenberg, and Sifa]{deusser2026evaluating}
Tobias Deu{\ss}er, Lorenz Sparrenberg, and Rafet Sifa.
\newblock {Evaluating and Benchmarking the System One Model Jev}.
\newblock \emph{arXiv preprint arXiv:2609.37647v1}, 2026.

\bibitem[Du et~al.(2026)Du, Liu, Zheng, Li, Guo, Zhang, and Zou]{du2026omnistreaming}
Enjun Du, Siyi Liu, Ziyu Zheng, Jingyu Li, Yiwen Guo, Yongqi Zhang, and Difan Zou.
\newblock Omni-streaming thinking.
\newblock \emph{arXiv preprint arXiv:2609.15128}, 2026.

\bibitem[Duan et~al.(2026)Duan, Lin, Wang, Xiang, Wu, Zhang, Yan, Wang, and Wei]{duan2026jeval}
Feiyu Duan, Jiayu Lin, Jia Wang, Jun Xiang, Jialiang Wu, Xinnong Zhang, Hanqi Yan, Siyuan Wang, and Zhongyu Wei.
\newblock {General Decision Models: Benchmarking and Insights Beyond Jev}.
\newblock \emph{arXiv preprint arXiv:2610.03935}, 2026.

\bibitem[Gneiting \& Raftery(2007)Gneiting and Raftery]{gneiting2007proper}
Tilmann Gneiting and Adrian~E. Raftery.
\newblock Strictly proper scoring rules, prediction, and estimation.
\newblock \emph{Journal of the American Statistical Association}, 102\penalty0 (477):\penalty0 359--378, 2007.

\bibitem[Guo et~al.(2017)Guo, Pleiss, Sun, and Weinberger]{guo2017calibration}
Chuan Guo, Geoff Pleiss, Yu~Sun, and Kilian~Q. Weinberger.
\newblock On calibration of modern neural networks.
\newblock In \emph{Proceedings of the 34th International Conference on Machine Learning}, pp.\  1321--1330, Sydney, Australia, 2017.

\bibitem[Guo et~al.(2026)Guo, Zhang, Guo, and Li]{guo2026gamewam}
Yuncheng Guo, Zhanqiu Zhang, Yiwen Guo, and Weijia Li.
\newblock {GameWAM}: A world action model for video games.
\newblock \emph{arXiv preprint arXiv:2608.26200}, 2026.

\bibitem[{H2O.ai}(2026)]{h2o2026lightning}
{H2O.ai}.
\newblock {H2O-Lightning-4B}, 2026.
\newblock URL \url{https://huggingface.co/h2oai/h2o-lightning-4b/blob/708e93926a034195d908e033fcb796c55398a6f5/README.md}.

\bibitem[Hafner(2022)]{hafner2022crafter}
Danijar Hafner.
\newblock Benchmarking the spectrum of agent capabilities.
\newblock In \emph{Proceedings of the 10th International Conference on Learning Representations}, pp.\  24538--24558, Virtual, 2022.

\bibitem[He et~al.(2024)He, Yao, Ma, Yu, Dai, Zhang, Lan, and Yu]{he2024webvoyager}
Hongliang He, Wenlin Yao, Kaixin Ma, Wenhao Yu, Yong Dai, Hongming Zhang, Zhenzhong Lan, and Dong Yu.
\newblock {WebVoyager}: Building an end-to-end web agent with large multimodal models.
\newblock In \emph{Proceedings of the 62nd Annual Meeting of the {A}ssociation for {C}omputational {L}inguistics (Volume 1: Long Papers)}, pp.\  6864--6890, Bangkok, Thailand, 2024.

\bibitem[He et~al.(2026)He, Chawla, Souri, Som, and Song]{he2026webstar}
Yifei He, Pranit Chawla, Yaser Souri, Subhojit Som, and Xia Song.
\newblock {WebSTAR}: Scalable data synthesis for computer use agents with step-level filtering.
\newblock In \emph{Proceedings of the 64th Annual Meeting of the Association for Computational Linguistics (Volume 1: Long Papers)}, pp.\  516--533, San Diego, CA, 2026.

\bibitem[Hu et~al.(2024)Hu, Shi, Fu, Roth, Ostendorf, Zettlemoyer, Smith, and Krishna]{hu2024visualsketch}
Yushi Hu, Weijia Shi, Xingyu Fu, Dan Roth, Mari Ostendorf, Luke Zettlemoyer, Noah~A. Smith, and Ranjay Krishna.
\newblock {Visual Sketchpad}: Sketching as a visual chain of thought for multimodal language models.
\newblock In \emph{Advances in Neural Information Processing Systems 37}, pp.\  139348--139379, Vancouver, Canada, 2024.

\bibitem[Huang et~al.(2026)Huang, Ding, Ju, Liu, Shah, and Zhao]{huang2026threshold}
Xingyue Huang, Xueying Ding, Mingxuan Ju, Yozen Liu, Neil Shah, and Tong Zhao.
\newblock Threshold differential attention for sink-free, ultra-sparse, and non-dispersive language modeling.
\newblock In \emph{Proceedings of the 64th Annual Meeting of the Association for Computational Linguistics (Volume 1: Long Papers)}, pp.\  18073--18092, San Diego, CA, 2026.

\bibitem[Kazemzadeh et~al.(2014)Kazemzadeh, Ordonez, Matten, and Berg]{kazemzadeh2014referitgame}
Sahar Kazemzadeh, Vicente Ordonez, Mark Matten, and Tamara Berg.
\newblock {ReferItGame}: Referring to objects in photographs of natural scenes.
\newblock In \emph{Proceedings of the 2014 Conference on Empirical Methods in Natural Language Processing}, pp.\  787--798, Doha, Qatar, 2014.

\bibitem[Kim et~al.(2025)Kim, Pertsch, Karamcheti, Xiao, Balakrishna, Nair, Rafailov, Foster, Sanketi, Vuong, Kollar, Burchfiel, Tedrake, Sadigh, Levine, Liang, and Finn]{kim2025openvla}
Moo~Jin Kim, Karl Pertsch, Siddharth Karamcheti, Ted Xiao, Ashwin Balakrishna, Suraj Nair, Rafael Rafailov, Ethan~P. Foster, Pannag~R. Sanketi, Quan Vuong, Thomas Kollar, Benjamin Burchfiel, Russ Tedrake, Dorsa Sadigh, Sergey Levine, Percy Liang, and Chelsea Finn.
\newblock {OpenVLA}: An open-source vision-language-action model.
\newblock In \emph{Proceedings of the 8th Conference on Robot Learning}, pp.\  2679--2713, Munich, Germany, 2025.

\bibitem[Koh et~al.(2024)Koh, Lo, Jang, Duvvur, Lim, Huang, Neubig, Zhou, Salakhutdinov, and Fried]{koh2024visualwebarena}
Jing~Yu Koh, Robert Lo, Lawrence Jang, Vikram Duvvur, Ming~Chong Lim, Po-Yu Huang, Graham Neubig, Shuyan Zhou, Ruslan Salakhutdinov, and Daniel Fried.
\newblock {VisualWebArena}: Evaluating multimodal agents on realistic visually grounded web tasks.
\newblock In \emph{Proceedings of the 62nd Annual Meeting of the {A}ssociation for {C}omputational {L}inguistics (Volume 1: Long Papers)}, pp.\  881--905, Bangkok, Thailand, 2024.

\bibitem[Lee et~al.(2019)Lee, Lee, Kim, Kosiorek, Choi, and Teh]{lee2019settransformer}
Juho Lee, Yoonho Lee, Jungtaek Kim, Adam~R. Kosiorek, Seungjin Choi, and Yee~Whye Teh.
\newblock {Set Transformer}: A framework for attention-based permutation-invariant neural networks.
\newblock In \emph{Proceedings of the 36th International Conference on Machine Learning}, pp.\  3744--3753, Long Beach, CA, 2019.

\bibitem[Lee(2026)]{theoleecj2026semif}
Theodore Lee.
\newblock {SemIf} (formerly {OpenJev}).
\newblock \url{https://github.com/TheoLeeCJ/SemIf-OpenJev}, 2026.

\bibitem[Li et~al.(2024)Li, Xie, Li, Chen, Wang, Chen, Yang, Wang, Kong, and Liu]{li2024vlfeedback}
Lei Li, Zhihui Xie, Mukai Li, Shunian Chen, Peiyi Wang, Liang Chen, Yazheng Yang, Benyou Wang, Lingpeng Kong, and Qi~Liu.
\newblock {VLFeedback}: A large-scale {AI} feedback dataset for large vision-language models alignment.
\newblock In \emph{Proceedings of the 2024 Conference on Empirical Methods in Natural Language Processing}, pp.\  6227--6246, Miami, FL, 2024.

\bibitem[Li et~al.(2025)Li, Wei, Xie, Yang, Song, Wang, An, Liu, Li, Lin, Kong, and Liu]{vlrewardbench}
Lei Li, Yuancheng Wei, Zhihui Xie, Xuqing Yang, Yifan Song, Peiyi Wang, Chenxin An, Tianyu Liu, Sujian Li, Bill~Yuchen Lin, Lingpeng Kong, and Qi~Liu.
\newblock {VL-RewardBench}: A challenging benchmark for vision-language generative reward models.
\newblock In \emph{Proceedings of the IEEE/CVF Conference on Computer Vision and Pattern Recognition}, pp.\  24657--24668, Nashville, TN, 2025.

\bibitem[Li et~al.(2026)Li, Lin, Wang, Zhang, Cui, Li, Wang, Liu, Chen, Bian, and Lam]{li2026forgewm}
Xinye Li, Lingshuai Lin, Lei Wang, Liuzhou Zhang, Jialin Cui, Qingshan Li, Guanchu Wang, Qingbin Liu, Xi~Chen, Jiang Bian, and Wai Lam.
\newblock {ForgeWM}: Progressive causal training for few-step action-conditioned video world models.
\newblock \emph{arXiv preprint arXiv:2608.14022}, 2026.

\bibitem[Li \& Wagle(2026)Li and Wagle]{li2026llmasjev}
Yinheng Li and Justin Wagle.
\newblock {LLM-as-Jev}: {LLMs} are already {Jev}-style decision models---when and how to fine-tune them.
\newblock \emph{arXiv preprint arXiv:2610.02076}, 2026.

\bibitem[Lin et~al.(2025)Lin, Li, Gao, Yang, Wu, Bai, Lei, Wang, and Shou]{lin2025showui}
Kevin~Qinghong Lin, Linjie Li, Difei Gao, Zhengyuan Yang, Shiwei Wu, Zechen Bai, Stan~Weixian Lei, Lijuan Wang, and Mike~Zheng Shou.
\newblock {ShowUI}: One vision-language-action model for {GUI} visual agent.
\newblock In \emph{Proceedings of the IEEE/CVF Conference on Computer Vision and Pattern Recognition}, pp.\  19498--19508, Nashville, TN, 2025.

\bibitem[Liu et~al.(2023)Liu, Li, Wu, and Lee]{liu2023llava}
Haotian Liu, Chunyuan Li, Qingyang Wu, and Yong~Jae Lee.
\newblock Visual instruction tuning.
\newblock In \emph{Advances in Neural Information Processing Systems 36}, pp.\  34892--34916, New Orleans, LA, 2023.

\bibitem[Lu et~al.(2022)Lu, Mishra, Xia, Qiu, Chang, Zhu, Tafjord, Clark, and Kalyan]{lu2022scienceqa}
Pan Lu, Swaroop Mishra, Tony Xia, Liang Qiu, Kai-Wei Chang, Song-Chun Zhu, Oyvind Tafjord, Peter Clark, and Ashwin Kalyan.
\newblock Learn to explain: Multimodal reasoning via thought chains for science question answering.
\newblock In \emph{Advances in Neural Information Processing Systems 35}, pp.\  2507--2521, New Orleans, LA, 2022.

\bibitem[Masry et~al.(2022)Masry, Long, Tan, Joty, and Hoque]{masry2022chartqa}
Ahmed Masry, Do~Xuan Long, Jia~Qing Tan, Shafiq Joty, and Enamul Hoque.
\newblock {ChartQA}: A benchmark for question answering about charts with visual and logical reasoning.
\newblock In \emph{Findings of the Association for Computational Linguistics: ACL 2022}, pp.\  2263--2279, Dublin, Ireland, 2022.

\bibitem[Mathew et~al.(2022)Mathew, Bagal, Tito, Karatzas, Valveny, and Jawahar]{infographicvqa}
Minesh Mathew, Viraj Bagal, Rub{\`e}n Tito, Dimosthenis Karatzas, Ernest Valveny, and C.V. Jawahar.
\newblock {InfographicVQA}.
\newblock In \emph{Proceedings of the IEEE/CVF Winter Conference on Applications of Computer Vision}, pp.\  1697--1706, Waikoloa, HI, 2022.

\bibitem[Meng et~al.(2025)Meng, Du, Liu, Zhou, Lu, Fu, Han, Shi, Wang, He, Zhang, Luo, Qiao, Zhang, and Shao]{meng2025mmeureka}
Fanqing Meng, Lingxiao Du, Zongkai Liu, Zhixiang Zhou, Quanfeng Lu, Daocheng Fu, Tiancheng Han, Botian Shi, Wenhai Wang, Junjun He, Kaipeng Zhang, Ping Luo, Yu~Qiao, Qiaosheng Zhang, and Wenqi Shao.
\newblock {MM-Eureka}: Exploring the frontiers of multimodal reasoning with rule-based reinforcement learning.
\newblock \emph{arXiv preprint arXiv:2503.07365}, 2025.

\bibitem[{Microsoft}(2026)]{microsoft2026decision}
{Microsoft}.
\newblock {Microsoft-Decision-1}, 2026.
\newblock URL \url{https://labs.ai.azure.com/innovations/microsoft-decision-1/}.

\bibitem[{OpenAI}(2024)]{openai2024gpt4o}
{OpenAI}.
\newblock {GPT-4o System Card}, 2024.
\newblock URL \url{https://openai.com/index/gpt-4o-system-card/}.

\bibitem[{OpenAI}(2026{\natexlab{a}})]{openai2026decisions}
{OpenAI}.
\newblock {Decisions}, 2026{\natexlab{a}}.
\newblock URL \url{https://developers.openai.com/api/docs/guides/decisions}.

\bibitem[{OpenAI}(2026{\natexlab{b}})]{openai2026gpt55}
{OpenAI}.
\newblock {Introducing GPT-5.5}, 2026{\natexlab{b}}.
\newblock URL \url{https://openai.com/index/introducing-gpt-5-5/}.

\bibitem[Ouyang et~al.(2026)Ouyang, Hu, Lin, Ng, and Shou]{gameworld}
Mingyu Ouyang, Siyuan Hu, Kevin~Qinghong Lin, Hwee~Tou Ng, and Mike~Zheng Shou.
\newblock {GameWorld}: Towards standardized and verifiable evaluation of multimodal game agents.
\newblock In \emph{Proceedings of the 19th European Conference on Computer Vision}, pp.\  41--59, Malm{\"o}, Sweden, 2026.

\bibitem[Patil et~al.(2025)Patil, Mao, Yan, Ji, Suresh, Stoica, and Gonzalez]{bfcl}
Shishir~G. Patil, Huanzhi Mao, Fanjia Yan, Charlie Cheng-Jie Ji, Vishnu Suresh, Ion Stoica, and Joseph~E. Gonzalez.
\newblock The {Berkeley Function Calling Leaderboard} ({BFCL}): From tool use to agentic evaluation of large language models.
\newblock In \emph{Proceedings of the 42nd International Conference on Machine Learning}, pp.\  48371--48392, Vancouver, Canada, 2025.

\bibitem[{Perplexity AI}(2026)]{perplexity2026decider}
{Perplexity AI}.
\newblock {pplx-decider-v1-27b}, 2026.
\newblock URL \url{https://huggingface.co/perplexity-ai/pplx-decider-v1-27b}.

\bibitem[{Qwen Team}(2025)]{qwen2025qwen3vl4b}
{Qwen Team}.
\newblock {Qwen3-VL-4B-Instruct}, 2025.
\newblock URL \url{https://huggingface.co/Qwen/Qwen3-VL-4B-Instruct}.

\bibitem[{Qwen Team}(2026{\natexlab{a}})]{qwen2026qwen35}
{Qwen Team}.
\newblock {Qwen3.5}, 2026{\natexlab{a}}.
\newblock URL \url{https://huggingface.co/collections/Qwen/qwen35}.

\bibitem[{Qwen Team}(2026{\natexlab{b}})]{qwen2026qwen36}
{Qwen Team}.
\newblock {Qwen3.6-27B}, 2026{\natexlab{b}}.
\newblock URL \url{https://huggingface.co/Qwen/Qwen3.6-27B}.

\bibitem[Ray et~al.(2025)Ray, Duan, Brown, Tan, Bashkirova, Hendrix, Ehsani, Kembhavi, Plummer, Krishna, Zeng, and Saenko]{ray2025sat}
Arijit Ray, Jiafei Duan, Ellis Brown, Reuben Tan, Dina Bashkirova, Rose Hendrix, Kiana Ehsani, Aniruddha Kembhavi, Bryan Plummer, Ranjay Krishna, Kuo-Hao Zeng, and Kate Saenko.
\newblock {SAT}: Dynamic spatial aptitude training for multimodal language models.
\newblock In \emph{Proceedings of the 2nd Conference on Language Modeling}, Montreal, Canada, 2025.

\bibitem[Robinson et~al.(2023)Robinson, Rytting, and Wingate]{robinson2023mcqa}
Joshua Robinson, Christopher~Michael Rytting, and David Wingate.
\newblock Leveraging large language models for multiple choice question answering.
\newblock In \emph{Proceedings of the 11th International Conference on Learning Representations}, pp.\  36470--36497, Kigali, Rwanda, 2023.

\bibitem[Ross et~al.(2025)Ross, Mahabaleshwarkar, and Suhara]{when2call}
Hayley Ross, Ameya~Sunil Mahabaleshwarkar, and Yoshi Suhara.
\newblock {W}hen2{C}all: When (not) to call tools.
\newblock In \emph{Proceedings of the 2025 Conference of the Nations of the {A}mericas Chapter of the {A}ssociation for {C}omputational {L}inguistics: Human Language Technologies (Volume 1: Long Papers)}, pp.\  3391--3409, Albuquerque, NM, 2025.

\bibitem[Schwenk et~al.(2022)Schwenk, Khandelwal, Clark, Marino, and Mottaghi]{schwenk2022aokvqa}
Dustin Schwenk, Apoorv Khandelwal, Christopher Clark, Kenneth Marino, and Roozbeh Mottaghi.
\newblock {A-OKVQA}: A benchmark for visual question answering using world knowledge.
\newblock In \emph{Proceedings of the 17th European Conference on Computer Vision}, pp.\  146--162, Tel Aviv, Israel, 2022.

\bibitem[Sermanet et~al.(2024)Sermanet, Ding, Zhao, Xia, Dwibedi, Gopalakrishnan, Chan, Dulac-Arnold, Maddineni, Joshi, Florence, Han, Baruch, Lu, Mirchandani, Xu, Sanketi, Hausman, Shafran, Ichter, and Cao]{sermanet2024robovqa}
Pierre Sermanet, Tianli Ding, Jeffrey Zhao, Fei Xia, Debidatta Dwibedi, Keerthana Gopalakrishnan, Christine Chan, Gabriel Dulac-Arnold, Sharath Maddineni, Nikhil~J. Joshi, Pete Florence, Wei Han, Robert Baruch, Yao Lu, Suvir Mirchandani, Peng Xu, Pannag Sanketi, Karol Hausman, Izhak Shafran, Brian Ichter, and Yuan Cao.
\newblock {RoboVQA}: Multimodal long-horizon reasoning for robotics.
\newblock In \emph{Proceedings of the IEEE International Conference on Robotics and Automation}, pp.\  645--652, Yokohama, Japan, 2024.

\bibitem[Shao et~al.(2024)Shao, Qian, Xiao, Song, Zong, Wang, Liu, and Li]{shao2024visualcot}
Hao Shao, Shengju Qian, Han Xiao, Guanglu Song, Zhuofan Zong, Letian Wang, Yu~Liu, and Hongsheng Li.
\newblock {Visual CoT}: Advancing multi-modal language models with a comprehensive dataset and benchmark for chain-of-thought reasoning.
\newblock In \emph{Advances in Neural Information Processing Systems 37}, pp.\  8612--8642, Vancouver, Canada, 2024.

\bibitem[Shen et~al.(2025{\natexlab{a}})Shen, Zhao, Zhao, Xu, Zhang, Zhu, and Yin]{shen2025zoomeye}
Haozhan Shen, Kangjia Zhao, Tiancheng Zhao, Ruochen Xu, Zilun Zhang, Mingwei Zhu, and Jianwei Yin.
\newblock {ZoomEye}: Enhancing multimodal {LLMs} with human-like zooming capabilities through tree-based image exploration.
\newblock In \emph{Proceedings of the 2025 Conference on Empirical Methods in Natural Language Processing}, pp.\  6602--6618, Suzhou, China, 2025{\natexlab{a}}.

\bibitem[Shen(2026)]{openjevdata2026}
Junhao Shen.
\newblock {OpenJevData-140k}, 2026.
\newblock URL \url{https://huggingface.co/datasets/shenjunhao/OpenJevData-140k}.

\bibitem[Shen et~al.(2025{\natexlab{b}})Shen, Zhao, Gu, Gao, Liu, Huang, Gao, Lin, Zhang, and Chen]{shen2025sophia}
Junhao Shen, Haiteng Zhao, Yuzhe Gu, Songyang Gao, Kuikun Liu, Haian Huang, Jianfei Gao, Dahua Lin, Wenwei Zhang, and Kai Chen.
\newblock Semi-off-policy reinforcement learning for vision-language slow-thinking reasoning.
\newblock In \emph{Advances in Neural Information Processing Systems 38}, pp.\  125082--125115, San Diego, CA and Mexico City, Mexico, 2025{\natexlab{b}}.

\bibitem[Shen et~al.(2026)Shen, Zong, Guo, Zhang, Guo, and Cheng]{mmdm2026}
Junhao Shen, Qing Zong, Yue Guo, Zhanqiu Zhang, Yiwen Guo, and Hong Cheng.
\newblock {MMDM}: Measuring massive multitask decision making.
\newblock \emph{arXiv preprint arXiv:2611.14309}, 2026.

\bibitem[Tan et~al.(2025)Tan, Zhang, Xu, Xia, Ding, Li, Zhou, Yue, Jiang, Li, An, Qin, Zong, Zheng, Wu, Chai, Bi, Xie, Gu, Li, Zhang, Tian, Wang, Wang, Karlsson, An, Yan, and Lu]{tan2025cradle}
Weihao Tan, Wentao Zhang, Xinrun Xu, Haochong Xia, Ziluo Ding, Boyu Li, Bohan Zhou, Junpeng Yue, Jiechuan Jiang, Yewen Li, Ruyi An, Molei Qin, Chuqiao Zong, Longtao Zheng, Yujie Wu, Xiaoqiang Chai, Yifei Bi, Tianbao Xie, Pengjie Gu, Xiyun Li, Ceyao Zhang, Long Tian, Chaojie Wang, Xinrun Wang, B{\"o}rje~F. Karlsson, Bo~An, Shuicheng Yan, and Zongqing Lu.
\newblock {Cradle}: Empowering foundation agents towards general computer control.
\newblock In \emph{Proceedings of the 42nd International Conference on Machine Learning}, pp.\  58658--58725, Vancouver, Canada, 2025.

\bibitem[Tong et~al.(2026)Tong, Tang, Li, Mou, Zhang, Zhao, Wen, Song, Zhan, Lu, Tao, Guo, Yu, Cheng, Xi, Jiang, Yin, Zheng, Ge, Chen, Gui, Qiu, Zhang, and Huang]{tong2026gamerl}
Jingqi Tong, Jixin Tang, Hangcheng Li, Yurong Mou, Ming Zhang, Jun Zhao, Yanbo Wen, Fan Song, Jiahao Zhan, Yuyang Lu, Chaoran Tao, Zhiyuan Guo, Jizhou Yu, Tianhao Cheng, Zhiheng Xi, Changhao Jiang, Zhangyue Yin, Yining Zheng, Weifeng Ge, Guanhua Chen, Tao Gui, Xipeng Qiu, Qi~Zhang, and Xuanjing Huang.
\newblock {Game-RL}: Synthesizing multimodal verifiable game data to boost {VLMs}' general reasoning.
\newblock In \emph{Proceedings of the 14th International Conference on Learning Representations}, pp.\  34689--34755, Rio de Janeiro, Brazil, 2026.

\bibitem[Wei et~al.(2025)Wei, Zhao, Sun, Lin, Yin, Hu, Zhang, Yu, Lv, Weng, Wang, Han, Ge, Zhang, Jiang, and Patel]{wei2025openvisionreasoner}
Yana Wei, Liang Zhao, Jianjian Sun, Kangheng Lin, Jisheng Yin, Jingcheng Hu, Yinmin Zhang, En~Yu, Haoran Lv, Zejia Weng, Jia Wang, Qi~Han, Zheng Ge, Xiangyu Zhang, Daxin Jiang, and Vishal~M. Patel.
\newblock {Open Vision Reasoner}: Transferring linguistic cognitive behavior for visual reasoning.
\newblock In \emph{Advances in Neural Information Processing Systems 38}, pp.\  4463--4494, San Diego, CA and Mexico City, Mexico, 2025.

\bibitem[{WeMM-JEV Team}(2026)]{wemmjevteam2026wemm}
{WeMM-JEV Team}.
\newblock {WeMM-JEV}, 2026.
\newblock URL \url{https://huggingface.co/Rover00/WeMM-JEV}.

\bibitem[Williams(1992)]{williams1992reinforce}
Ronald~J. Williams.
\newblock Simple statistical gradient-following algorithms for connectionist reinforcement learning.
\newblock \emph{Machine Learning}, 8\penalty0 (3--4):\penalty0 229--256, 1992.

\bibitem[Wu \& Xie(2024)Wu and Xie]{wu2024vstar}
Penghao Wu and Saining Xie.
\newblock V$*$: Guided visual search as a core mechanism in multimodal {LLMs}.
\newblock In \emph{Proceedings of the IEEE/CVF Conference on Computer Vision and Pattern Recognition}, pp.\  13084--13094, Seattle, WA, 2024.

\bibitem[Xu et~al.(2026)Xu, Du, Gao, Yang, Shen, Chen, Yao, Guo, Lu, Huang, Lu, Liu, Zheng, Zhang, Zou, and Liu]{xu2026workagents}
Zishan Xu, Enjun Du, Mingqi Gao, Cheng Yang, Junhao Shen, Zirong Chen, Zhiyuan Yao, Yifu Guo, Yuquan Lu, Jinyang Huang, Zhengxi Lu, Siyi Liu, Ziyu Zheng, Yongqi Zhang, Difan Zou, and Weiwen Liu.
\newblock Work agents: Systems, capabilities, and optimization.
\newblock \emph{arXiv preprint arXiv:2610.35845}, 2026.

\bibitem[Yang et~al.(2025)Yang, Chen, Zhang, Zhao, Qian, Wang, Wang, Koripella, Movahedi, Li, Ji, Zhang, and Zhang]{embodiedbench}
Rui Yang, Hanyang Chen, Junyu Zhang, Mark Zhao, Cheng Qian, Kangrui Wang, Qineng Wang, Teja~Venkat Koripella, Marziyeh Movahedi, Manling Li, Heng Ji, Huan Zhang, and Tong Zhang.
\newblock {EmbodiedBench}: Comprehensive benchmarking multi-modal large language models for vision-driven embodied agents.
\newblock In \emph{Proceedings of the 42nd International Conference on Machine Learning}, pp.\  70576--70631, Vancouver, Canada, 2025.

\bibitem[Yao et~al.(2023)Yao, Zhao, Yu, Du, Shafran, Narasimhan, and Cao]{yao2023react}
Shunyu Yao, Jeffrey Zhao, Dian Yu, Nan Du, Izhak Shafran, Karthik~R. Narasimhan, and Yuan Cao.
\newblock {ReAct}: Synergizing reasoning and acting in language models.
\newblock In \emph{Proceedings of the 11th International Conference on Learning Representations}, pp.\  30084--30116, Kigali, Rwanda, 2023.

\bibitem[Yu et~al.(2016)Yu, Poirson, Yang, Berg, and Berg]{yu2016refcoco}
Licheng Yu, Patrick Poirson, Shan Yang, Alexander~C. Berg, and Tamara~L. Berg.
\newblock Modeling context in referring expressions.
\newblock In \emph{Proceedings of the 14th European Conference on Computer Vision}, pp.\  69--85, Amsterdam, Netherlands, 2016.

\bibitem[Yu et~al.(2025)Yu, Zhang, Li, Xu, Yao, Chen, Lu, Cui, Dang, He, Feng, Song, Zheng, Liu, Chua, and Sun]{yu2025rlaifv}
Tianyu Yu, Haoye Zhang, Qiming Li, Qixin Xu, Yuan Yao, Da~Chen, Xiaoman Lu, Ganqu Cui, Yunkai Dang, Taiwen He, Xiaocheng Feng, Jun Song, Bo~Zheng, Zhiyuan Liu, Tat-Seng Chua, and Maosong Sun.
\newblock {RLAIF-V}: Open-source {AI} feedback leads to super {GPT-4V} trustworthiness.
\newblock In \emph{Proceedings of the IEEE/CVF Conference on Computer Vision and Pattern Recognition}, pp.\  19985--19995, Nashville, TN, 2025.

\bibitem[Yuan et~al.(2026)Yuan, Cui, Huang, Chen, Ni, Dong, Li, Zheng, Tang, and Hao]{yuan2026embodiedr1}
Yifu Yuan, Haiqin Cui, Yaoting Huang, Yibin Chen, Fei Ni, Zibin Dong, Pengyi Li, Yan Zheng, Hongyao Tang, and Jianye Hao.
\newblock {Embodied-R1}: Reinforced embodied reasoning for general robotic manipulation.
\newblock In \emph{Proceedings of the 14th International Conference on Learning Representations}, Rio de Janeiro, Brazil, 2026.

\bibitem[Yue et~al.(2025)Yue, Zheng, Ni, Wang, Zhang, Tong, Sun, Yu, Zhang, Sun, Su, Chen, and Neubig]{mmmupro}
Xiang Yue, Tianyu Zheng, Yuansheng Ni, Yubo Wang, Kai Zhang, Shengbang Tong, Yuxuan Sun, Botao Yu, Ge~Zhang, Huan Sun, Yu~Su, Wenhu Chen, and Graham Neubig.
\newblock {MMMU-Pro}: A more robust multi-discipline multimodal understanding benchmark.
\newblock In \emph{Proceedings of the 63rd Annual Meeting of the {A}ssociation for {C}omputational {L}inguistics (Volume 1: Long Papers)}, pp.\  15134--15186, Vienna, Austria, 2025.

\bibitem[Zhang et~al.(2025)Zhang, Huang, Yao, Liu, Zhang, Lu, and Tao]{zhang2025r1vl}
Jingyi Zhang, Jiaxing Huang, Huanjin Yao, Shunyu Liu, Xikun Zhang, Shijian Lu, and Dacheng Tao.
\newblock {R1-VL}: Learning to reason with multimodal large language models via step-wise group relative policy optimization.
\newblock In \emph{Proceedings of the IEEE/CVF International Conference on Computer Vision}, pp.\  1859--1869, Honolulu, HI, 2025.

\bibitem[Zhang et~al.(2020)Zhang, Walshe, Liu, Guan, Muller, Whritner, Zhang, Hayhoe, and Ballard]{zhang2020atarihead}
Ruohan Zhang, Calen Walshe, Zhuode Liu, Lin Guan, Karl~S. Muller, Jake~A. Whritner, Luxin Zhang, Mary~M. Hayhoe, and Dana~H. Ballard.
\newblock {Atari-HEAD}: {Atari} human eye-tracking and demonstration dataset.
\newblock In \emph{Proceedings of the 34th AAAI Conference on Artificial Intelligence}, pp.\  6811--6820, New York, NY, 2020.

\bibitem[Zheng et~al.(2024)Zheng, Zhou, Meng, Zhou, and Huang]{zheng2024selectors}
Chujie Zheng, Hao Zhou, Fandong Meng, Jie Zhou, and Minlie Huang.
\newblock Large language models are not robust multiple choice selectors.
\newblock In \emph{Proceedings of the 12th International Conference on Learning Representations}, pp.\  19426--19454, Vienna, Austria, 2024.

\bibitem[Zhu et~al.(2026{\natexlab{a}})Zhu, Xu, He, Chai, Hu, Zheng, Luo, Li, Zhang, and Cheng]{zhu2026intern}
Tong Zhu, Zhe Xu, Zefeng He, Yunkai Chai, Xuyang Hu, Dehong Zheng, Yun Luo, Yafu Li, Qi~Zhang, and Yu~Cheng.
\newblock {Intern-Decision}: Efficient multimodal decision making with autoregressive masked language modeling.
\newblock \emph{arXiv preprint arXiv:2610.22131}, 2026{\natexlab{a}}.

\bibitem[Zhu et~al.(2026{\natexlab{b}})Zhu, Zeng, Si, Yang, Li, Qin, Zhou, Yao, Zhang, Qin, Zhang, Yang, Xu, Zhang, Guo, Wang, Duan, and Wang]{zhu2026videojev}
Yuhan Zhu, Xiangyu Zeng, Qingyi Si, Chenxu Yang, Yifei Li, Luozheng Qin, Junhao Zhou, Dingyu Yao, Jun Zhang, Chuanyu Qin, Zhiqiu Zhang, Yuandong Yang, Yuqi Xu, Tianyu Zhang, Hongcheng Guo, Jiaqi Wang, Nan Duan, and Limin Wang.
\newblock {VideoJev}: System one model for video understanding and interaction.
\newblock \emph{arXiv preprint arXiv:2610.18615}, 2026{\natexlab{b}}.

\end{thebibliography}
\endgroup

\clearpage
\begingroup
  \renewcommand{\familydefault}{\sfdefault}
  \sffamily\fontsize{10.8}{16}\selectfont
  \renewcommand{\sectionfont}{\fontsize{15.5}{19}\selectfont}
  \setlength{\parskip}{0pt}
  \renewcommand{\contentsname}{Appendix Contents}
  \pdfbookmark[1]{Appendix Contents}{appendix-contents}
  \tableofcontents
\endgroup
\clearpage
\beginappendix
\addtocontents{toc}{\protect\setcounter{tocdepth}{2}}
\suppressfloats[t]
\setcounter{figure}{0}
\setcounter{table}{0}
\setcounter{equation}{0}
\setcounter{algorithm}{0}
\renewcommand{\thefigure}{S\arabic{figure}}
\renewcommand{\thetable}{S\arabic{table}}
\renewcommand{\theequation}{S\arabic{equation}}
\renewcommand{\thealgorithm}{S\arabic{algorithm}}
\renewcommand{\theHfigure}{app.\arabic{figure}}
\renewcommand{\theHtable}{app.\arabic{table}}
\renewcommand{\theHequation}{app.\arabic{equation}}
\AddToHook{cmd/section/before}{\Needspace{5\baselineskip}}
\section{Architecture and Learning Objective}
\label{app:derivations}
\subsection{Model Configuration}
Table~\ref{tab:arch} lists the backbone and decision-head dimensions for both models.
The 4B and 27B models project backbone states of width 2,560 and 5,120, respectively, into a common 512-dimensional decision space.
Both heads use two set-attention blocks followed by a bilinear readout.
Each block contains an eight-head attention layer and a two-layer feed-forward network with GELU. Both sublayers use pre-normalization and a residual connection.
The decision head operates in FP32 on BF16 backbone states and normalizes its scores over the valid options.
Both training stages update the language backbone and decision head and keep the vision encoder and merger frozen.

\subsection{Permutation Equivariance}
\label{app:equivariance}
Consider a decision $x=(s,q,\optionSet)$ with $K$ options. Let $\bm{P}\in\{0,1\}^{K\times K}$ be a permutation matrix and $x_{\bm{P}}=(s,q,\optionSet_{\bm{P}})$ the same decision with reordered options. Each option retains its name, criterion, and references to other options, while the validity mask is reordered accordingly. We show that
\begin{equation}
  \decisionmodel(x_{\bm{P}})=\bm{P}\decisionmodel(x).
  \label{eq:equivariance}
\end{equation}

Since all options share the same state and query prefix but are encoded in separate branches, reordering them only changes the order of their hidden representations. Adding the same projected query to each option preserves this property.
In the set-attention blocks, all options share the same projection weights, and no option-position embeddings are used. Thus, reordering the options transforms the attention score matrix $\bm{A}\in\mathbb{R}^{K\times K}$ into $\bm{P}\bm{A}\bm{P}^{\top}$. Row-wise softmax satisfies
$$
  \operatorname{softmax}(\bm{P}\bm{A}\bm{P}^{\top})
  =\bm{P}\operatorname{softmax}(\bm{A})\bm{P}^{\top}.
$$
Therefore, the attention outputs follow the same permutation. The remaining operations, including normalization, feed-forward layers, residual connections, and the query-conditioned readout, also preserve this property. The final softmax thus produces the reordered probability distribution, proving Eq.~\eqref{eq:equivariance}.
Consequently, reordering options does not change their assigned probabilities.
\begin{table}[t]
\reporttable
\caption{\textbf{Configurations of the released models.} Each hybrid group is three Gated DeltaNet layers and one full-attention layer. Token budgets are 32,768 per branch and 65,536 per decision.}
\label{tab:arch}
\begin{tabularx}{\linewidth}{lYY}
\toprule
\tablehead{Component} & \tablehead{4B} & \tablehead{27B} \\
\midrule
\rowcolor{tablegroupbg}\multicolumn{3}{l}{\tablehead{Backbone (BF16)}} \\
Layers (hybrid groups $N$) & 32 (8) & 64 (16) \\
Hidden size $H$ & 2,560 & 5,120 \\
Attention heads (Q / KV) & 16 / 4 & 24 / 4 \\
DeltaNet heads (K / V) & 16 / 32 & 16 / 48 \\
MLP width & 9,216 & 17,408 \\
Vision layers (frozen) & 24 & 27 \\
Parameters (with vision) & 4.54B & 26.09B \\
\midrule
\rowcolor{tablegroupbg}\multicolumn{3}{l}{\tablehead{Decision head (FP32)}} \\
$D$ / blocks / heads & \multicolumn{2}{c}{512 / 2 / 8} \\
Set FFN width & \multicolumn{2}{c}{2,048} \\
Parameters & 9.45M & 12.07M \\
\bottomrule
\end{tabularx}
\end{table}

\subsection{The RLCD Objective and Its Gradient}
\label{app:proper}
Let $\bm{p}$ and $\targetdist$ denote the predicted and target distributions over $K$ options. From Eq.~\eqref{eq:population}, the difference between the predicted loss and the loss at the target distribution is
\begin{align}
 &\ell_\alpha(\bm{p},\targetdist)-\ell_\alpha(\targetdist,\targetdist)\notag\\
 &\quad=\tfrac{\alpha}{2}\|\bm{p}-\targetdist\|_2^2+
                 (1-\alpha)D_{\mathrm{KL}}(\targetdist\|\bm{p}).
 \label{eq:proper}
\end{align}

Both terms are nonnegative, and their sum is zero only when $\bm{p}=\targetdist$. Thus, the loss is minimized when the prediction matches the target distribution, even if some target probabilities are zero (using $0\log 0=0$).
For one-hot targets, softmax predictions can approach this minimum as the logits grow.
The additional KL penalty in Eq.~\eqref{eq:rl} limits how far the updated model moves from the frozen SFT policy.

As discussed in Section~\ref{sec:training}, annotations can be noisy, and a single recorded choice does not tell us how likely each option is.
Thus, RLCD records whether a sampled option matches the observed outcome. For a given option, the probability of a match is its target probability.
To learn the outcome distribution from binary feedback, we separate the part of the gradient that requires an observed outcome from the part we can compute in a single forward pass.
For a fixed decision, let $r_i$ and $p_i$ denote the target and predicted probabilities of option $i$.
Differentiating the probability loss gives
\begin{align}
 \nabla_{\bm{\theta}}\ell_\alpha
 &=\nabla_{\bm{\theta}}\tfrac{\alpha}{2}\sum_i p_i^2 \notag\\
 &\quad-\sum_i r_i(\alpha p_i+1-\alpha)
                        \nabla_{\bm{\theta}}\log p_i .
 \label{eq:exact-gradient}
\end{align}
The first term depends only on the $K$ predicted probabilities.
We sum it exactly, along with the gradient of the reference penalty, so neither contribution incurs sampling variance.
Only the second term requires feedback.
The feedback source draws one outcome $y\sim\mathrm{Cat}(\targetdist)$ for the decision.
Independently of $y$, we draw $G$ option indices $a_g$ with replacement from the proposal $\mu_i=(1-\epsilon)p_i+\epsilon/K$.
Each draw returns a match bit $c_g=\mathbf{1}[a_g=y]$.
With the importance weight included in $k_g=(\alpha p_{a_g}+1-\alpha)c_g/\mu_{a_g}$, the score-function estimate~\citep{williams1992reinforce} satisfies
\begin{align}
 \mathbb E_{a_g\sim\bm{\mu}}[k_g\nabla_{\bm{\theta}}\log p_{a_g}\mid y]
      =(\alpha p_y+1-\alpha)\nabla_{\bm{\theta}}\log p_y .
\end{align}
Averaging over $y$ recovers the target-weighted sum in Eq.~\eqref{eq:exact-gradient}. All $G$ draws share that outcome and the same model prediction.

For $G>1$, let $b_g=\sum_{h\ne g}k_h/(G-1)$ and $w_g=p_{a_g}/\mu_{a_g}$. Given $y$, the leave-one-out baseline $b_g$ is independent of $a_g$. Its weighted contribution has zero expectation because
\begin{align}
 \mathbb E[w_g b_g\nabla_{\bm{\theta}}\log p_{a_g}\mid y]
 =\mathbb E[b_g\mid y]\underbrace{\sum_i\nabla_{\bm{\theta}} p_i}_{=\,\bm{0}}.
\end{align}
Subtracting $w_g b_g$ from $k_g$ therefore preserves the expected gradient.
Algorithm~\ref{alg:rlcd} holds these coefficients fixed when differentiating the sampled log probabilities, giving the unbiased estimator in Eq.~\eqref{eq:rl}.
Uniform exploration ensures $\mu_i\geq\epsilon/K$ and, for $\epsilon<1$, $w_g\leq 1/(1-\epsilon)$.

The unbiasedness result above is conditional on the training target $\targetdist$. For vote-based soft labels, $r_i=n_i/N$ is the fraction of $N$ annotators who choose option $i$. Vote counts of $(1,2)$ and $(10,20)$ give the same target, although 30 votes give a more precise estimate of the underlying preference probabilities than three when votes are independent draws from the same annotator population.
RLCD fits the empirical vote shares. Once these shares are fixed, $N$ does not enter the estimator separately. Thus, both sets of vote counts define the same loss for this decision, even though a confidence interval for the population preference would depend on the number of votes.

The \emph{soft gradient gate} operates on the feedback term in each update. Given outcome $y$, the probability that all $G$ option draws miss $y$ is $(1-\mu_y)^G$. In that case, all match indicators and leave-one-out baselines for this decision are zero, and only the analytic terms contribute to its gradient.
When one or more matches occur, importance weighting corrects for the option-sampling distribution. At fixed model parameters, averaging over outcomes and option draws recovers the full gradient of the empirical objective. Even a vote share that is unusually high or low by chance retains its weight in this expectation. Here, \emph{filtering} refers specifically to the stochastic gating of outcome-dependent gradient contributions within an individual update.
Under the fixed training budget in Appendix~\ref{app:additional}, SFT is more sample-efficient on hard-label decisions, while assigning most soft-label examples to RLCD yields lower Brier scores.

\section{Training Data and Supervision}
\label{app:data}
This section details how we construct, filter, and allocate the training data introduced in Sections~\ref{sec:data} and~\ref{sec:training}.
Each training example consists of a state $s$, a query $q$, and a set of options $\optionSet$, together with a target probability distribution over these options.
Visual observations are included in the state or relevant option descriptions, while source labels and verification results are used to construct and validate the targets without being exposed to the model.

\subsection{Constructing Decisions}
\noindent\textbf{Annotated Text.}
Although LightDecision-1 is a multimodal model, high-quality text data remain essential for learning how to make structured decisions. We build on OpenJevData-140k~\citep{openjevdata2026}, which provides diverse decision examples in English and Chinese, covering question answering, logical reasoning, numerical tasks, and preference judgments.
The dataset also includes compositional problems constructed from facts, rules, and option descriptions, with challenging distractors that require reasoning over multiple constraints. These examples are validated through rule-based checks, executable solutions, or independent model judgments, and provide both hard and soft labels.
Together, they help the model learn to interpret queries, compare options, and assign meaningful probabilities across different decision tasks.

\noindent\textbf{Visual Evidence and Actions.}
Image questions come from ScienceQA~\citep{lu2022scienceqa}, A-OKVQA~\citep{schwenk2022aokvqa}, and ChartQA~\citep{masry2022chartqa}.
We also use the SAT~\citep{ray2025sat} and RefCOCO~\citep{kazemzadeh2014referitgame,yu2016refcoco} spatial annotations distributed with Embodied-R1~\citep{yuan2026embodiedr1}.
RLAIF-V~\citep{yu2025rlaifv} and VLFeedback~\citep{li2024vlfeedback} supply response preferences, and WebSTAR~\citep{he2026webstar} supplies judgments of interface trajectories.
For action prediction, we pair the observations available at decision time with the recorded next action, as in EgoPlan-IT~\citep{chen2026egoplan} and the human demonstrations in Crafter~\citep{hafner2022crafter}.
A recorded action tells us what the agent did. To determine which action works better, we use execution outcomes. In Procgen~\citep{cobbe2020procgen}, for example, we run each candidate plan from the same saved state and rank the plans by completion, failure, and reward.

\subsection{Curation and Stage Assignment}
\label{app:curation}
\noindent\textbf{Filtering Visual Shortcuts.}
We remove textual cues that reveal the answer. In ScienceQA we keep only questions with an image and omit the worked solution. In RoboVQA~\citep{sermanet2024robovqa} we strip the answer from the source query before asking whether the observed task is complete or an action is feasible.
We also construct options that cannot be distinguished by matching their wording to the query. In RefCOCO, for example, the query names an object, but all four options are coordinates. We keep a decision only if exactly one point lies inside the object's segmentation mask. The model sees the unmarked image.
For visual preferences, we discard empty or identical responses and pairs with extreme length differences. In VLFeedback we require the preferred response to score at least two points higher on visual faithfulness and at least as high on helpfulness.

\noindent\textbf{Label Checks.}
We verify that each target assigns finite, nonnegative probabilities to the intended options and sums to one. We merge numerically equivalent options and discard duplicate decisions with conflicting labels.
For next-action prediction, the observation history ends before the labeled action. We exclude EgoPlan decisions with a missing cutoff frame and restrict the action history to steps completed by that frame. For reward comparisons on Atari-HEAD~\citep{zhang2020atarihead}, we discard windows with repeated or nearly static frames and check the labels against recorded rewards.

\noindent\textbf{Separating Training from Evaluation.}
We reserve GameWorld tasks and the ALFRED and Habitat episodes in EmbodiedBench for evaluation~\citep{gameworld,embodiedbench}. Training uses separate game recordings from Crafter, Atari-HEAD, and GameWAM~\citep{guo2026gamewam}, independently seeded MiniGrid~\citep{chevalierboisvert2023minigrid} and Procgen environments, and embodied observations from RoboVQA and EgoPlan.
Before splitting, we group samples from the same source video, trajectory, or seeded world and join groups that share an identical decoded image. For RoboVQA, the original sequence record is the finest reliable trajectory identifier.
We remove GameQA~\citep{tong2026gamerl} training examples with the same game and board state as an upstream test example, and EgoPlan windows that overlap upstream validation windows.
We also screen indexed VL-RewardBench and AgentRewardBench images for exact and near duplicates. For resynthesized text decisions, we filter exact and high-overlap matches to the source questions and check for overlap with MMDM.
Visual SFT examples include image questions and response preferences. RLCD introduces decisions from game and robot trajectories.
As summarized in Table~\ref{tab:data}, RLCD also revisits selected text and visual examples from SFT, applying the same objective to new and replayed contexts.

\begin{table}[t]
\reporttable
\caption{\textbf{Data allocation for post-training.} Replay reuses SFT examples and is included in the total RL pool.}
\label{tab:data}
\begin{tabularx}{\linewidth}{lrX}
\toprule
\tablehead{Training pool} & \tablehead{Approx. size} & \tablehead{Role} \\
\midrule
SFT & 116K & Learn the decision interface across text and visual inputs. \\
Primary RL & 97K & Refine decision distributions and introduce game and embodied tasks. \\
SFT replay in RL & 29K & Revisit selected SFT contexts under the RLCD objective. \\
\midrule
Total RL & 126K & Combine primary contexts with sample replay. \\
\bottomrule
\end{tabularx}
\end{table}

\section{Evaluation Protocols}
\label{app:protocol}
\begin{table*}[t]
\reporttable
\caption{\textbf{Evaluation sets.} Sizes count the scored unit of each metric, questions unless a unit is given. DI denotes the Decision Index and AA denotes Artificial Analysis.}
\label{tab:benchmarks}
\begin{tabularx}{\linewidth}{*{4}{>{\raggedright\arraybackslash}X}}
\toprule
\tablehead{Benchmark} & \tablehead{Subset} & \tablehead{Evaluation size} & \tablehead{Primary metric} \\
\midrule
\rowcolor{tablegroupbg}\multicolumn{4}{c}{\tablehead{Visual and interactive benchmarks}} \\
GameWorld & 34 games & 170 tasks & Progress and success rate \\
MMMU-Pro & Vision, test & 1,730 & Accuracy \\
InfographicVQA & DI public MCQ-244 & 244 & Accuracy \\
EmbodiedBench-ALFRED & Six subsets & 300 tasks & Success rate \\
EmbodiedBench-Habitat & Six subsets & 300 tasks & Success rate \\
VL-RewardBench & Test & 1,247 pairs & Three-domain macro accuracy \\
\midrule
\rowcolor{tablegroupbg}\multicolumn{4}{c}{\tablehead{Text benchmarks}} \\
MMDM & Single-select & 4,112 & Accuracy \\
MMDM & Multi-select & 633 & Exact-set accuracy \\
MMDM & Soft labels & 703 & Full-distribution Brier \\
When2Call & AA-filtered & 3,423 & Accuracy \\
BFCL & Tool selection & 1,694 cases & Case exact accuracy \\
\bottomrule
\end{tabularx}
\end{table*}

Table~\ref{tab:benchmarks} lists the evaluation sets and their denominators. Appendix~\ref{app:baselines} describes the model groups and the mapping from each model output to a benchmark prediction.

\subsection{Interaction Protocols}
\label{app:interaction}
GameWorld pauses during inference and supplies the current screenshot (RGB JPEG, longest side 768 pixels), public task goal, valid controls, and up to two previous rounds. The model first chooses a semantic action and then any discrete parameters required, such as a board cell. The evaluator runs each task once, up to 100 environment steps, and observes the simulator's private state only to compute the progress and completion measures.
For EmbodiedBench, ALFRED and Habitat each contain 300 episodes across six equally sized high-level capability subsets. We follow the official step and invalid-action budgets and provide the observation and available high-level actions at each step. Task success is determined by the terminal environment flag.

\subsection{Visual Question Answering and Tool Selection}
\label{app:di}
We follow the Decision Index 0.3 construction and scoring protocols~\citep{decisionindex} for BFCL, MMMU-Pro vision, and InfographicVQA. On MMMU-Pro, we use the vision subset and measure accuracy against reference answers. InfographicVQA provides four options, with distractors drawn from other questions about the same infographic. Images in both visual benchmarks use a consistent RGB format, with the longest side limited to 1,568 pixels and the shortest side padded to at least 280 pixels. On BFCL~\citep{bfcl}, a prediction is correct only when the selected functions exactly match the reference set, and we evaluate 1,694 cases.
For When2Call~\citep{when2call}, we use the 3,423 questions selected by Artificial Analysis. Each question has four options in a fixed, per-question permutation. For WeMM-JEV-27B, we report the \href{https://huggingface.co/datasets/Rover00/WeMM-JEV-Results/blob/main/runs/wemm-jev-text/benchmark-summary.json}{authors\textquoteright{} Decision Index 0.3 results} on 3,652 When2Call questions and 1,694 BFCL cases.

We evaluate \methodname{} 4B and 27B, Perplexity Decider-V1, and H2O-Lightning-4B on all three benchmarks, and Jev on the two visual benchmarks. The GPT-4o run covers 1,716 of the 1,730 MMMU-Pro vision questions. Its accuracy uses all 1,730 questions as the denominator. We also include results from the public Decision Index leaderboard, marked with $\dagger$ in the result tables. For Clef-Omni and Intern-Decision-4B, the leaderboard reports chance-adjusted skill scores $s_{\mathrm{DI}}$ of Decision Index 0.3 on MMMU-Pro vision and InfographicVQA. We convert these scores to accuracy using $a_{\mathrm{chance}}+s_{\mathrm{DI}}(1-a_{\mathrm{chance}})$, where $a_{\mathrm{chance}}$ is 0.1281 for MMMU-Pro and 0.25 for InfographicVQA. We then round to the corresponding number of correct answers.

\subsection{MMDM Benchmark}
\label{app:limits}
We report three metrics from MMDM~\citep{mmdm2026}.
\textbf{Hard Single-select Accuracy} is the fraction of questions on which the model ranks the correct option above every other option.
\textbf{Hard Multi-select Accuracy} is the fraction on which every correct option ranks above every incorrect option.
\textbf{Soft Brier} measures how closely the predicted probabilities match the reference distribution over $N_{\mathrm{eval}}$ soft-label questions. For question $n$ with $K_n$ options, let $p_{n,i}$ and $r_{n,i}$ denote the predicted and reference probabilities of option $i$, respectively. The score is
\begin{equation}
  \operatorname{Brier}=\frac{1}{N_{\mathrm{eval}}}\sum_{n=1}^{N_{\mathrm{eval}}}
             \sum_{i=1}^{K_n}(p_{n,i}-r_{n,i})^2 .
  \label{eq:brier}
\end{equation}
Lower scores indicate a closer match.

\subsection{Inference Latency}
For each locally timed model and readout in Table~\ref{tab:vision}, we report the 95th percentile of per-question inference times in seconds. We exclude API-based models from latency comparisons because their latency depends on external factors that cannot be controlled in a fair comparison.
We process one MMMU-Pro vision question at a time on a single GPU with approximately one BF16 petaFLOP/s of peak throughput. Each model encodes the image and prefills the shared context once per question. Decision baselines score the options in a single prompt. For Qwen models, the letter-probability readout uses the same prompt layout and normalizes next-token probabilities over the answer letters. Autoregressive baselines generate the answer token by token with reasoning disabled. \methodname{} scores the option branches in parallel from the cached context.
We synchronize device timing to include image encoding, prefill, and all forward passes through the final decision. Input preprocessing, host-to-device transfer, and answer parsing are excluded.
\section{Analysis Protocols}
\label{app:analysis-protocols}
\subsection{Visual Probability Evaluation}
\label{app:visual-probability}
Figure~\ref{fig:visual-calibration} uses the same 1,730 MMMU-Pro vision questions and 244 four-choice InfographicVQA questions as Table~\ref{tab:vision}. We align the returned option keys with the reference and require a finite, normalized probability vector for every question. \methodname{} uses its native head at softmax temperature $T=1$. Qwen probabilities are conditional on the supplied answer letters. WeMM-JEV-27B uses its released probability readout.
For each model and benchmark, we group questions by the probability assigned to the selected option using ten equal-width bins on $[0,1]$, with probability one included in the final bin~\citep{guo2017calibration}.
For each nonempty bin, we plot mean confidence on the horizontal axis and the fraction of correct selections on the vertical axis.

\subsection{Matched-State Case Studies}
Figure~\ref{fig:interaction} pairs recorded trajectories of \methodname{} 27B and WeMM-JEV-27B on two tasks that only \methodname{} completes. For each pair, we locate the first step at which the models choose different actions. Both see the same image, query, ordered options, and history up to that choice. Each bar shows the probability assigned to that action over the full option set. The right-hand frames come from each model's own trajectory after three actions, counting the first differing action. The episode outcomes come from the terminal records.

\subsection{Permutation Equivariance}
\label{app:order-protocol}
We sort the 244 four-choice InfographicVQA question IDs lexicographically and run full-pipeline rotations on the first 48. For each, repeat the original request and evaluate the three other cyclic orders with the same model-visible names, descriptions, and image. For each cyclic permutation matrix $\bm{P}\in\{0,1\}^{4\times4}$, let $\bm{p}_{\mathrm{rot}},\bm{p}\in\Delta^{3}$ be the probability vectors for the rotated and original lists, respectively. We restore the original order and compute $\|\bm{P}^{\top}\bm{p}_{\mathrm{rot}}-\bm{p}\|_\infty$. For each model, we compare each rotated prediction with its original-order prediction. The flip rate divides the number of changed top choices by 144. The affected-question count includes each question once if any of its three rotations changes the top choice. 

\subsection{Measuring the Width of Executable Action Sets}
\label{app:action-width}
Figure~\ref{fig:action-set-success} groups 170 GameWorld tasks, 300 ALFRED episodes, and 300 Habitat episodes by the number of options available for a single decision. In ALFRED and Habitat, the action menu is fixed within an episode, so its size gives the episode's option count. In GameWorld, the menu can change from step to step. We use the largest number of options offered in any one control or parameter decision along the \methodname{} 27B reference trajectory. Matching by task ID assigns the same width to all four models.
We pool the three benchmarks and group tasks into the inclusive ranges 1--77, 78--199, 200--249, and 250--305 options. These bins contain 465, 152, 106, and 47 tasks, respectively. The first bin ends at the largest option count used in training. Within each bin, a model's SR is the percentage of tasks it successfully completes. Each task contributes one binary outcome regardless of episode length, with missing outcomes counted as failures as in Table~\ref{tab:vision}.

\subsection{Controlled Ablation Settings}
\label{app:attribution}
\noindent\textbf{Joint Option Modeling.}
The three heads use the same training data and hyperparameters. The \emph{independent bilinear readout} scores each projected option vector directly. The \emph{self-only attention} variant first refines each vector through both set-attention blocks, including their feed-forward layers, with attention restricted to the diagonal. The \emph{full model} allows each option to attend to all valid options in both blocks. Comparing the latter two variants isolates cross-option interaction while keeping the depth and feed-forward layers of the head the same. All three use a shared state and query prefix and normalize the option scores together at $T=1$.

\noindent\textbf{Post-Training Strategy.}
\emph{SFT only} evaluates the checkpoint at the end of the first stage. The three second-stage variants start from this checkpoint and share the training data, optimizer settings, and KL penalty to the frozen SFT policy. \emph{Continued cross-entropy} fits the full target distribution with the SFT loss. The \emph{full-target quadratic--log loss} also uses the full target and differentiates Eq.~\eqref{eq:population} exactly. RLCD uses the same quadratic--log objective but estimates its outcome-dependent gradient from binary matches. Thus, the last two variants keep the scoring rule fixed while changing how the target enters the update.

\noindent\textbf{Effectiveness of RLCD.}
From the same SFT checkpoint, we compare three updates under matched training settings. \emph{Full sampling} estimates both the prediction-only and outcome-dependent parts of the scoring gradient from sampled options. \emph{PPO with ratio clipping} reuses sampled feedback across updates and uses a clipped policy ratio in its objective. RLCD draws $G=16$ options with replacement per decision, with 10\% uniform exploration. Each draw is compared with the same observed outcome. RLCD computes the prediction-only and KL gradients exactly and applies importance weighting and a leave-one-out baseline to the sampled outcome term (Algorithm~\ref{alg:rlcd}).

\section{Baseline Models and Settings}
\label{app:baselines}
\noindent\textbf{Proprietary Models.}
We evaluate Typesafe jev-1.14.0, Cloudflare Clef-Omni, Perplexity Decider-V1 via the OpenRouter Decisions API. We evaluate GPT-6 Luna Decisions and the general-purpose baselines GPT-4o (2024-08-06 snapshot) and GPT-5.5 via its official API. To fairly compare GPT-5.5 with decision models, we set \texttt{reasoning\_effort} to \texttt{none}.

\noindent\textbf{Open-Source Backbones.}
At 4B, the general-purpose baselines are Qwen3-VL-4B-Instruct and Qwen3.5-4B. The decision baselines are Intern-Decision-4B and H2O-Lightning-4B. Intern-Decision-4B uses its released calibration, and H2O-Lightning uses the scoring configuration distributed with its checkpoint. At 27B, we evaluate Qwen3.5-27B and Qwen3.6-27B alongside WeMM-JEV-27B. Each Qwen backbone has two readouts in Table~\ref{tab:vision}. For rows marked $^{\ddagger}$, we divide the next-token probability of each answer letter by the total probability assigned to the supplied answer letters. Rows marked $^{*}$ map the autoregressively generated answer to an option. Both use the same weights with reasoning disabled. WeMM-JEV-27B uses its released decision engine with permutation averaging disabled. 

\noindent\textbf{Option Limits and Score Sources.}
The interactive evaluations give the models the same observations and available actions. When an action list exceeds the native limit for Intern-Decision-4B (62 options) or Jev (255), we score groups of options and compare their winners until a single action remains, which is used in MMDM~\citep{mmdm2026}. This procedure yields an action for the simulator. Appendix~\ref{app:protocol} gives run coverage and identifies the visual knowledge scores recovered from the Decision Index leaderboard.

\section{Decoding Configurations}
\label{app:prompt-decoding}

\noindent\textbf{Native decision probabilities.}
Given a state, a query, and a set of options, both sizes of \methodname{} score all options in a single forward pass and apply a softmax at temperature $T=1$. For a single-choice question or an environment action, we select the option with the highest probability. For calibration, we retain the full distribution.

\noindent\textbf{Answer-letter probabilities ($^{\ddagger}$).}
For the Qwen baselines, we also read the next-token probabilities at the answer position. Let $x$ denote the input and $\ell_i$ the answer-letter token for option $i$. We normalize over the letters assigned to the $K$ available options, giving $p_i=P(\ell_i\mid x)/\sum_{j=1}^{K}P(\ell_j\mid x)$. This yields an option distribution in one forward pass. We use the highest-probability option for single-choice evaluation and the full distribution for calibration.

\noindent\textbf{Autoregressive selection ($^{*}$).}
General-purpose vision-language models generate a direct answer with reasoning disabled. The evaluator maps the answer to one of the options in the request. When a simulator requires a vector, the adapter encodes the selected action as a one-hot vector. This encoding records the action taken and is excluded from probability calibration.

\section{Additional Results}
\label{app:additional}

\subsection{Sample Efficiency of SFT and RLCD}
Given a fixed set of training examples, which labels are most useful at each stage? This question is important because training data for decision models, especially soft-labeled examples, are relatively scarce and should be used efficiently. It also helps us better understand how SFT and RLCD benefit from different types of supervision. We investigate this through a series of experiments.
We draw pools of 25,600 examples from the training data and compare cross-entropy SFT with RLCD. We draw an additional 500 hard-label questions and 1,000 soft-label questions, hold them out of training, and use them only to evaluate the allocations in this section. Both stages update the Qwen3.5-4B language backbone and the two-layer permutation-equivariant set-attention head while keeping the vision encoder frozen. All runs use seed 42 and the optimization settings in Section~\ref{sec:infrastructure}. In the hard-only and soft-only experiments, each run sees the same previously unseen examples in the same order. Only the point at which training switches from SFT to RLCD changes.

\begin{figure*}[t]
  \centering
  \includegraphics[width=\textwidth]{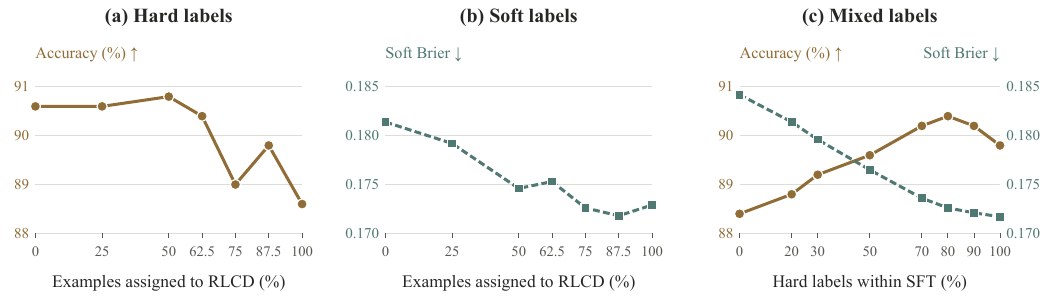}
  \caption{\textbf{Allocating a fixed training budget between SFT and RLCD.}
  In (a) and (b), the horizontal axis gives the fraction of examples assigned to RLCD. The rest go to SFT. In (c), each stage receives 12,800 examples, and their hard-label shares sum to 100\%. Circles show accuracy on 500 hard-label decisions and squares show mean Brier score on 1,000 soft-label decisions. Panel (c) uses the left axis for accuracy and the right for Brier.}
  \label{fig:sample-efficiency}
\end{figure*}

\noindent\textbf{Hard labels.}
As shown in Figure~\ref{fig:sample-efficiency}a, selection accuracy remains high and stable as long as SFT receives at least half of the hard-label pool, whereas shifting most or all hard examples to RLCD noticeably degrades performance. Because cross-entropy deterministically pulls logits toward one-hot targets without sampling variance, SFT is more sample-efficient for establishing the decision format and crisp selection boundaries. In contrast, relying solely on stochastic binary feedback in RLCD requires more samples to sharpen hard-label choices.

\noindent\textbf{Soft labels.}
Figure~\ref{fig:sample-efficiency}b shows the opposite trend.
Soft Brier error decreases steadily as soft examples move from SFT to RLCD, reaching its minimum when RLCD receives 87.5\% of the pool and remaining consistently low above 75\%. Unlike cross-entropy alone, RLCD's strictly proper quadratic--log scoring rule directly aligns predicted probabilities with target distributions, while a small SFT warm-up provides a better-aligned initialization than running RLCD from scratch.

\noindent\textbf{Exchanging labels between stages.}
We fix both stage budgets at 12,800 examples and use the same 12,800 hard and 12,800 soft examples in every run. Each hard example moved into SFT exchanges places with a soft example in RLCD. Raising the hard-label share in SFT from 0\% to 80\% improves both accuracy (88.4\% to 90.4\%) and Brier (0.1842 to 0.1726). Moving the remaining hard labels into SFT lowers Brier but costs more correct decision (Figure~\ref{fig:sample-efficiency}c). At this budget, a predominantly hard-label SFT stage and a predominantly soft-label RLCD stage give the highest observed accuracy while retaining most of the gain in probability quality.
 This result supports the label allocation used in training \methodname{}.

\subsection{Text Decision Benchmarks}
\label{app:text}
\begin{table*}[t]
\reporttable
\caption{\textbf{Text decision results on MMDM and tool use.} \ranklegend{} Accuracy is reported as a percentage; Brier uses its original scale. $^{\dagger}$~Published Decision Index results (Appendix~\ref{app:protocol}). The WeMM-JEV-27B When2Call score covers 3,652 questions, compared with 3,423 in our evaluation.}
\label{tab:text}
\begin{tabularx}{\linewidth}{l*{5}{Y}}
\toprule
\multirow{2}{*}{\tablehead{Model}} & \multicolumn{3}{c}{\tablehead{MMDM}} & \multicolumn{2}{c}{\tablehead{Tools}} \\
\cmidrule(lr){2-4}\cmidrule(lr){5-6}
 & \tablehead{Single-select$\uparrow$} & \tablehead{Multi-select$\uparrow$} & \tablehead{Soft Brier$\downarrow$} & \tablehead{When2Call$\uparrow$} & \tablehead{BFCL$\uparrow$} \\
\midrule
\rowcolor{tablegroupbg}\multicolumn{6}{c}{\tablehead{Proprietary Models}} \\
Jev 1.14 & 67.17 & 22.12 & \secondscore{0.1101} & 80.31 & 96.40\rlap{$^\dagger$} \\
Perplexity Decider-V1 & 62.38 & 21.48 & 0.1555 & 78.64 & \secondscore{97.99} \\
Clef-Omni & 56.91 & 14.85 & 0.2761 & 61.88 & \bestscore{98.47} \\
\midrule
\rowcolor{tablegroupbg}\multicolumn{6}{c}{\tablehead{Open-source Models}} \\
Intern-Decision-4B & 49.03 & 11.06 & 0.1564 & 68.80 & 92.74\rlap{$^\dagger$} \\
H2O-Lightning-4B & 58.29 & 20.54 & 0.3482 & 82.03 & 93.51 \\
WeMM-JEV-27B & \secondscore{76.00} & \bestscore{36.81} & 0.2327 & \secondscore{84.50}\rlap{$^\dagger$} & 97.93\rlap{$^\dagger$} \\
\rowcolor{tableoursbg}\textbf{LightDecision-1 4B} & 60.72 & 20.06 & 0.3442 & 73.04 & 85.89 \\
\rowcolor{tableoursbg}\textbf{LightDecision-1 27B} & \bestscore{76.41} & \secondscore{36.33} & \bestscore{0.0565} & \bestscore{85.07} & 95.87 \\
\bottomrule
\end{tabularx}
\end{table*}

The same decision head processes textual states when no image is supplied. Table~\ref{tab:text} evaluates general decisions on MMDM~\citep{mmdm2026} and tool selection on When2Call~\citep{when2call} and BFCL~\citep{bfcl}, using the model groups and readout notation of the main comparison. MMDM contains 4,112 single-select questions, 633 multi-select questions, and 703 soft-label questions. We report accuracy separately for the two hard-label subsets and Brier score for soft labels. When2Call measures option accuracy; BFCL requires the selected function set to match the reference. Appendix~\ref{app:protocol} gives the scoring rules.
Overall, \methodname{} 27B retains strong pure-text capability, ranking first on MMDM single-select, soft Brier, and When2Call, while \methodname{} 4B leads all 4B baselines on MMDM single-select.
More importantly, comparing hard-label and soft-label tracks reveals a clear calibration advantage.
Although \methodname{} 27B and WeMM-JEV-27B perform within half a point on both hard-label MMDM tracks, \methodname{} 27B achieves roughly four times lower soft Brier error, demonstrating that similar top-1 accuracy can hide stark differences in probability calibration. On BFCL, moving from 4B to 27B reduces \methodname{}'s exact-set errors from 239 to 70. The ranking across models also depends on the task. Clef-Omni makes only 26 errors on BFCL, the fewest in Table~\ref{tab:text}, yet ranks last on When2Call.

\subsection{Generalization to Larger Option Sets}
\label{app:sets}
The base training recipe uses at most 77 options per decision, and our visual and interactive evaluations reach 305. Table~\ref{tab:large-options} extends the comparison to MMDM questions with up to 1,024 options. 
Across all four widths, \methodname{} 27B ranks first consistently, while most baselines collapse near zero once $K \ge 256$ due to the attention dilution in single-sequence scoring. Besides, accuracy does not drop monotonically with $K$ because the $K=256$ split is dominated by strict multi-select items (18 of 26), where misranking a single distractor invalidates the entire prediction.

\begin{table*}[t]
\reporttable
\caption{\textbf{Accuracy on MMDM items with large option sets.} Counts give correct answers out of all questions. \ranklegend{} Failures remain in the denominator. Both LightDecision models return valid predictions on every question. Their request budget is 65,536 tokens for the first three groups and 131,072 tokens for 1,024 options.}
\label{tab:large-options}
\begin{tabularx}{\linewidth}{l*{4}{Y}}
\toprule
\tablehead{Model} & \tablehead{$K=255$} & \tablehead{$K=256$} & \tablehead{$K=512$} & \tablehead{$K=1{,}024$} \\
\midrule
\rowcolor{tableoursbg}\textbf{LightDecision-1 4B} & \secondscore{41.67} (15/36) & \secondscore{7.69} (2/26) & 33.33 (14/42) & 7.69 (2/26) \\
\rowcolor{tableoursbg}\textbf{LightDecision-1 27B} & \bestscore{66.67} (24/36) & \bestscore{38.46} (10/26) & \bestscore{54.76} (23/42) & \bestscore{38.46} (10/26) \\
Jev 1.14 & \secondscore{41.67} (15/36) & \secondscore{7.69} (2/26) & \secondscore{42.86} (18/42) & \secondscore{23.08} (6/26) \\
Perplexity Decider-V1 & 0.00 (0/36) & 0.00 (0/26) & 0.00 (0/42) & 0.00 (0/26) \\
Intern-Decision-4B & 19.44 (7/36) & 0.00 (0/26) & 2.38 (1/42) & 0.00 (0/26) \\
H2O-Lightning-4B & 22.22 (8/36) & 0.00 (0/26) & 4.76 (2/42) & 0.00 (0/26) \\
\bottomrule
\end{tabularx}
\end{table*}

\section{Qualitative Examples}
\label{app:examples}
\begingroup
\definecolor{caseink}{HTML}{566069}
\definecolor{casegold}{HTML}{96712B}
\definecolor{casefour}{HTML}{C6A75F}
\definecolor{casegreen}{HTML}{386957}
\newcommand{\casehead}[3]{%
  \par\noindent{\sffamily\bfseries\normalsize
    \textcolor{casegold}{#1}\hspace{.7em}#2}\hfill{\small\color{caseink}#3}\par\nobreak}
\newcommand{\casestills}[2]{%
  \par\vspace{3pt}\noindent\includegraphics[width=\linewidth]{figures/examples/#1}\par\nobreak
  \vspace{2pt}\noindent #2\par}
\newcommand{\casestep}[3]{%
  \begin{minipage}[t]{.1945\linewidth}\centering
    {\footnotesize\sffamily\bfseries #1\;\textperiodcentered\;#2\par}%
    {\footnotesize\color{caseink}#3\par}%
  \end{minipage}}
\newcommand{\casenote}[1]{\vspace{4pt}\noindent{\small #1}\par}
\newcommand{\caserule}{\par\vspace{9pt}\noindent{\color{black!18}\rule{\linewidth}{.4pt}}\par\vspace{8pt}}
\newlength{\casebarwidth}
\setlength{\casebarwidth}{14mm}
\newcommand{\quizbar}[3]{%
  \makebox[\casebarwidth][l]{\rlap{\color{black!7}\rule{\casebarwidth}{.75ex}}%
    {\color{#1}\rule{\dimexpr\casebarwidth*#2/1000\relax}{.75ex}}}%
  \hspace{.35em}\makebox[3em][r]{#3}}
\setlength{\parindent}{0pt}
\setlength{\parskip}{2pt}

\subsection*{MMMU-Pro Vision}

\vspace{4pt}
\casehead{01}{Recognizing a painting}{\href{https://huggingface.co/datasets/MMMU/MMMU_Pro/viewer/vision/test?row=78}{Art $\cdot$ Test 87}}
What is the title of this painting?

\vspace{4pt}
\noindent\begin{minipage}[c]{.28\linewidth}
  \centering\includegraphics[width=.95\linewidth]{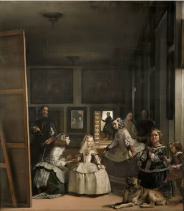}
\end{minipage}\hfill
\begin{minipage}[c]{.69\linewidth}
  \small
  \setlength{\tabcolsep}{3pt}
  \renewcommand{\arraystretch}{1.32}
  \begin{tabular*}{\linewidth}{@{\extracolsep{\fill}}lrr@{}}
    \textbf{Option} & \textcolor{casefour}{\textbf{4B}} & \textcolor{casegold}{\textbf{27B}}\\
    A. Floral Still Life & \quizbar{casefour}{1}{0.1\%} & \quizbar{casegold}{0}{$<$0.1\%}\\
    B. The Musicians & \quizbar{casefour}{1}{$<$0.1\%} & \quizbar{casegold}{0}{$<$0.1\%}\\
    C. The Scream & \quizbar{casefour}{0}{$<$0.1\%} & \quizbar{casegold}{0}{$<$0.1\%}\\
    D. The Night Watch & \quizbar{casefour}{1}{0.1\%} & \quizbar{casegold}{1}{$<$0.1\%}\\
    \textcolor{casegreen}{\textbf{E. Las Meninas}} & \quizbar{casefour}{996}{99.6\%} & \quizbar{casegold}{999}{99.9\%}\\
    F. The Liver is the Cock's Comb & \quizbar{casefour}{0}{$<$0.1\%} & \quizbar{casegold}{0}{$<$0.1\%}\\
    G. The Starry Night & \quizbar{casefour}{0}{$<$0.1\%} & \quizbar{casegold}{0}{$<$0.1\%}\\
    H. The Girl with a Pearl Earring & \quizbar{casefour}{0}{$<$0.1\%} & \quizbar{casegold}{0}{$<$0.1\%}\\
    I. The Last Supper & \quizbar{casefour}{0}{$<$0.1\%} & \quizbar{casegold}{0}{$<$0.1\%}\\
    J. Mona Lisa & \quizbar{casefour}{0}{$<$0.1\%} & \quizbar{casegold}{0}{$<$0.1\%}\\
  \end{tabular*}
\end{minipage}
\casenote{Both models identify \emph{Las Meninas}, assigning it more than 99\% of the probability across ten paintings.}

\caserule
\casehead{02}{Checking a musical transposition}{\href{https://huggingface.co/datasets/MMMU/MMMU_Pro/viewer/vision/test?row=71}{Music $\cdot$ Test 53}}
The second bar attempts to transpose the first down a minor third to sounding pitch. Is its third note correct?

\begin{center}
  \includegraphics[width=.97\linewidth]{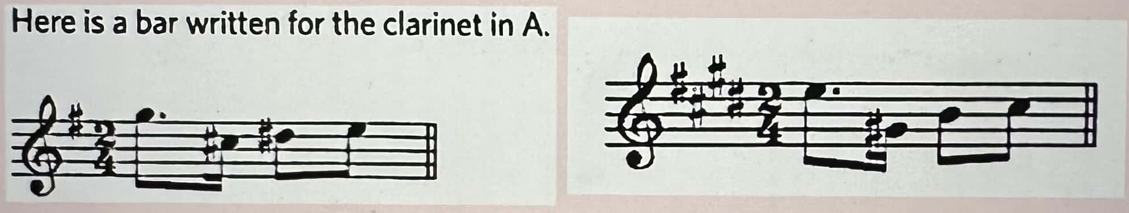}
\end{center}
{\small\setlength{\casebarwidth}{39mm}\setlength{\tabcolsep}{5pt}\renewcommand{\arraystretch}{1.55}
  \begin{tabular*}{\linewidth}{@{\extracolsep{\fill}}lrr@{}}
    \textbf{Option} & \textcolor{casefour}{\textbf{4B}} & \textcolor{casegold}{\textbf{27B}}\\
    A. True & \quizbar{casefour}{351}{35.1\%} & \quizbar{casegold}{217}{21.7\%}\\
    \textcolor{casegreen}{\textbf{B. False}} & \quizbar{casefour}{649}{64.9\%} & \quizbar{casegold}{783}{78.3\%}\\
  \end{tabular*}\par}
\casenote{Both models flag the third note as incorrect. The 27B model assigns 78.3\% to this answer, compared with 64.9\% for the 4B model.}

\clearpage
\subsection*{GameWorld}

\vspace{4pt}
\casehead{03}{2048}{Task 01\_04}
Merge tiles to reach a total score of 512.
\casestills{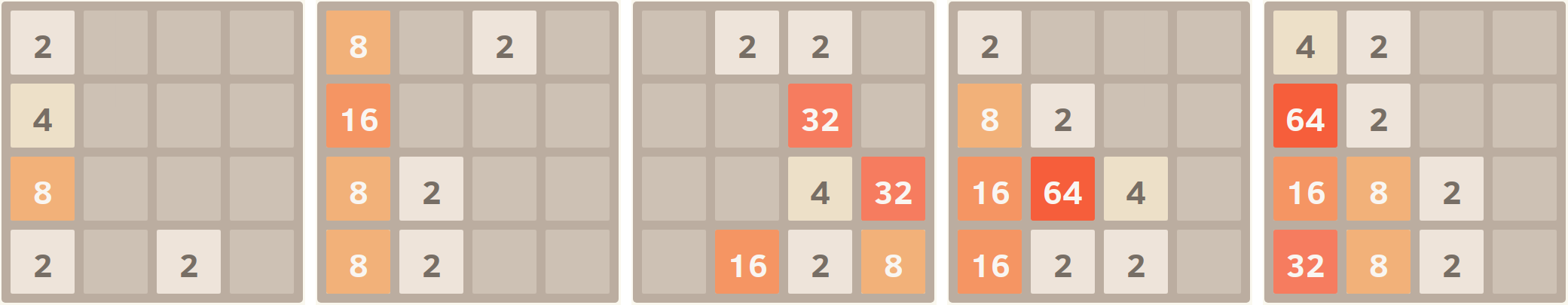}{%
  \casestep{9}{Left}{24 points}\hfill
  \casestep{25}{Down}{116 points}\hfill
  \casestep{49}{Up}{304 points}\hfill
  \casestep{57}{Down}{444 points}\hfill
  \casestep{69}{Up}{528 points}}
\casenote{The final upward move takes the score from 504 to 528, crossing the target on step 69.}

\caserule
\casehead{04}{Google Snake}{Task 15\_02}
Eat three apples in one run.
\casestills{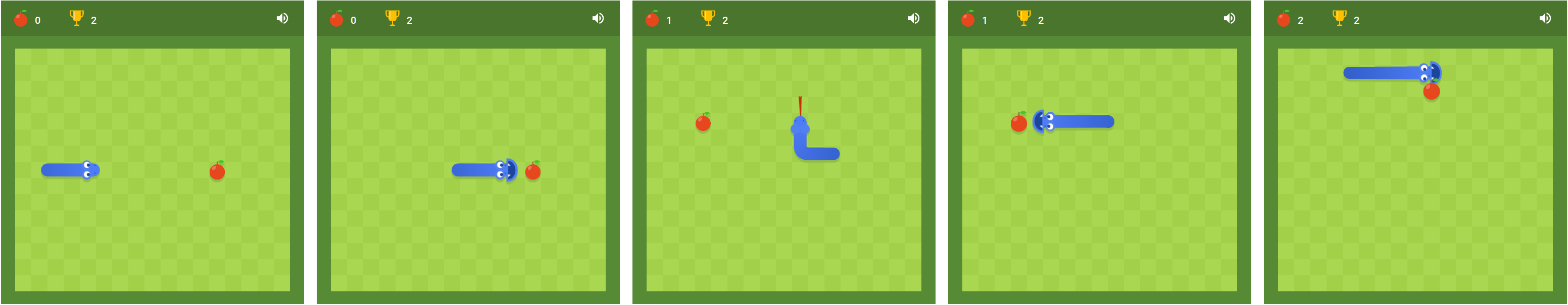}{%
  \casestep{24}{Right}{0 apples}\hfill
  \casestep{28}{Wait}{1 apple}\hfill
  \casestep{33}{Left}{1 apple}\hfill
  \casestep{36}{Left}{2 apples}\hfill
  \casestep{43}{Down}{3 apples}}
\casenote{In the final run, the snake reaches the first apple without changing direction. Left and downward inputs take it to the next two.}

\caserule
\casehead{05}{Pac-Man}{Task 23\_01}
Reach 600 points by clearing pellets in the maze.
\casestills{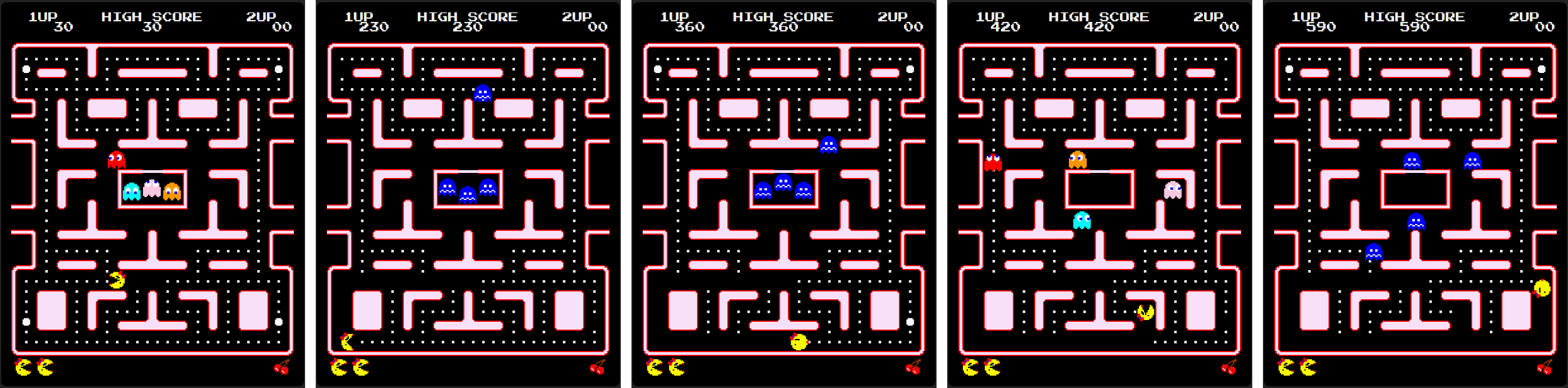}{%
  \casestep{6}{Left}{60 points}\hfill
  \casestep{11}{Right}{260 points}\hfill
  \casestep{15}{Right}{390 points}\hfill
  \casestep{27}{Down}{420 points}\hfill
  \casestep{33}{Up}{600 points}}
\casenote{The last six moves add 180 points. The final upward move reaches the target.}

\clearpage
\subsection*{EmbodiedBench}

\vspace{4pt}
\casehead{06}{Moving two books from the bed to the desk}{ALFRED $\cdot$ Base 031}
\casestills{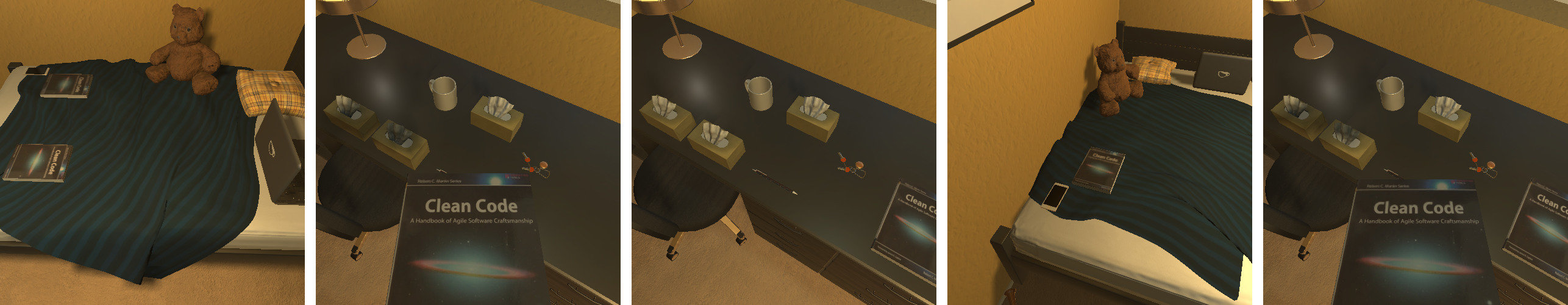}{%
  \casestep{2}{Pick up book}{39.1\%}\hfill
  \casestep{4}{Place book}{53.4\%}\hfill
  \casestep{5}{Find bed}{41.3\%}\hfill
  \casestep{7}{Pick up book}{55.9\%}\hfill
  \casestep{9}{Place book}{80.4\%}}
\casenote{After placing the first book, the model returns to the bed. Among 187 options, it ranks finding the bed (41.3\%) above searching directly for the second book (24.5\%). It delivers both books in nine actions.}

\caserule
\casehead{07}{Slicing and chilling lettuce}{ALFRED $\cdot$ Long horizon 026}
Cut a slice of lettuce, chill it, and put it in the sink.
\casestills{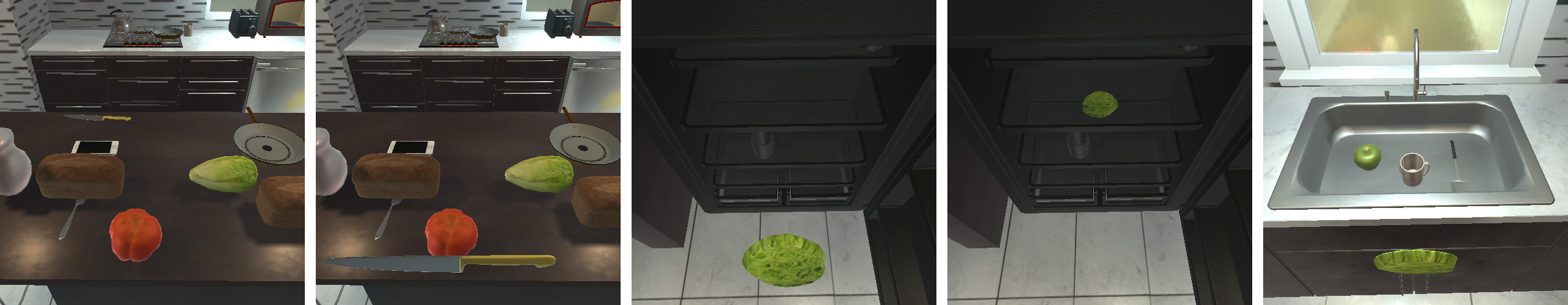}{%
  \casestep{7}{Pick up knife}{45.7\%}\hfill
  \casestep{8}{Slice lettuce}{22.9\%}\hfill
  \casestep{14}{Place in fridge}{38.6\%}\hfill
  \casestep{20}{Pick up lettuce}{93.9\%}\hfill
  \casestep{23}{Place in sink}{48.7\%}}
\casenote{With the knife in hand at step 8, slicing ranks first among 248 options, ahead of picking up the lettuce (14.0\%). After refrigeration, picking up the lettuce becomes the top choice.}

\caserule
\casehead{08}{Swapping a drill and a plum}{Habitat $\cdot$ Base 028}
Exchange the objects on the sofa and in the sink.
\casestills{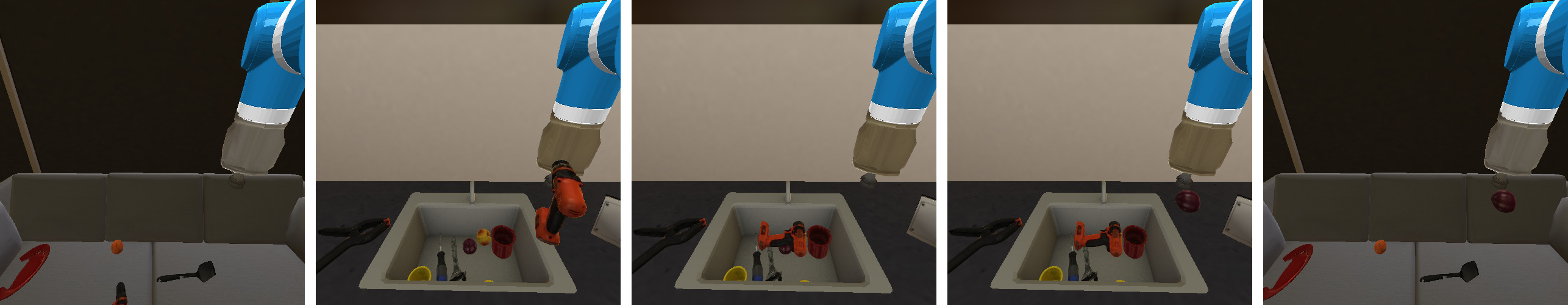}{%
  \casestep{2}{Pick up drill}{Sofa}\hfill
  \casestep{4}{Place drill}{Sink}\hfill
  \casestep{5}{Pick up plum}{Sink}\hfill
  \casestep{6}{Go to sofa}{Carrying plum}\hfill
  \casestep{7}{Place plum}{Sofa}}
\casenote{The sink is the destination for the drill and the pickup location for the plum. The model completes the exchange in seven actions.}
\endgroup

\end{document}